\documentclass[10pt,twocolumn,letterpaper]{article}

\usepackage{iccv}
\usepackage{times}
\usepackage{epsfig}
\usepackage{graphicx}
\usepackage{amsmath}
\usepackage{amssymb}

\usepackage{multirow}
\usepackage{makecell}
\usepackage[font=small]{caption}
\usepackage{overpic}

\usepackage[utf8]{inputenc}
\usepackage{amsthm}
\usepackage{booktabs}
\usepackage{enumitem}
\usepackage{pifont}
\usepackage{subcaption}
\usepackage{color}
\usepackage{diagbox}
\usepackage{array}
\usepackage{overpic}
\usepackage[table]{xcolor}

\makeindex

\graphicspath{{./figs/}}

\usepackage{pifont}
\usepackage{overpic}
\usepackage{url}

\newcommand{\nameofatten}{Re-attention}

\newcommand{\addFig}[1]{}
\newcommand{\addFigs}[1]{}

\definecolor{mygreen}{RGB}{0,100,0}
\definecolor{myred}{RGB}{150,0,0}
\definecolor{myblue}{RGB}{60,60,250}

\newcommand{\myPara}[1]{\vspace{.05in}\noindent\textbf{#1}}

\usepackage{hyperref}
\hypersetup{pagebackref=false,colorlinks=true,linkcolor=myred,citecolor=myblue,bookmarks=false,urlcolor=black}

\iccvfinalcopy % *** Uncomment this line for the final submission

\def\iccvPaperID{4046} % *** Enter the ICCV Paper ID here

\ificcvfinal\fi

\begin{document}

%%%%%%%%% TITLE
\title{Depthwise Scalable Vision Transformer on ImageNet-1k Without Extra Dataset}
\title{Does a Deep Vision Transformer perform Better?}
\title{Two Vision Transformers Blocks are Larger Than One? }
\title{DeepViT: Towards Depth-Scalable Vision Transformer}
\title{DeepViT: Towards Deeper Vision Transformer}

% \and

\author{Daquan Zhou\textsuperscript{1},
Bingyi Kang\textsuperscript{1},
Xiaojie Jin\textsuperscript{2},
Linjie Yang\textsuperscript{2},\\
Xiaochen Lian\textsuperscript{2},
Zihang Jiang\textsuperscript{1},
Qibin Hou\textsuperscript{1},
Jiashi Feng\textsuperscript{1} \\
\textsuperscript{1}National University of Singapore,\quad \textsuperscript{2}ByteDance US AI Lab  \\
\texttt{\small {\{zhoudaquan21, xjjin0731, lianxiaochen, yljatthu, andrewhoux\}}@gmail.com}
\\
\texttt{\small{jzihang, kang, elefjia}@nus.edu.sg}
}

\maketitle
\ificcvfinal\thispagestyle{empty}\fi

%%%%%%%%% ABSTRACT
\begin{abstract}
Vision transformers (ViTs) have been successfully applied in image classification tasks recently.
In this paper, we show that, unlike convolution neural networks (CNNs) that can be improved by
stacking more convolutional layers, the performance of ViTs saturate fast when scaled to be deeper.
More specifically, we empirically observe that such scaling difficulty is caused by the \emph{attention collapse} issue: as the transformer goes deeper, 
the attention maps gradually become similar and even much the same after certain layers. 
% \textcolor{red}{[JS: can we firmly say similar attention leads to similar features?]} 
In other words, the feature maps tend to be identical in the top layers of
deep ViT models. This fact demonstrates that in deeper layers of ViTs, the self-attention mechanism fails to learn effective concepts for representation learning and hinders the model from getting expected performance gain.
Based on above observation, we propose a simple yet effective method, named \emph{\nameofatten{}}, to re-generate the attention maps to increase their diversity at different layers with negligible computation and memory cost. 
The proposed method makes it feasible to train deeper ViT models with consistent performance improvements
via minor modification to existing ViT models.
Notably, when training a deep ViT model with 32 transformer blocks, the Top-1 classification
accuracy  can be improved by 1.6\% on ImageNet. 
Code is publicly available at \url{https://github.com/zhoudaquan/dvit_repo}.

%   and find that   the over-smoothing issue  of the self-attention maps as the model goes deeper is an important reason. Specifically, we find that as the transformer goes deeper, the attention maps from  different layers becomes increasingly more similar. This indicates that in deeper layers, the self-attention  does not contribute new information for representation learning and hinder    the performance gain. We then propose a simple yet effective method to re-generate the attention maps to increase their diversity at different layers. With this method, we can successfully train a ViT model as deep as xx layers, with monotonically performance improvement. The Top-1 classification accuracy of deep vision transformers can be improved by 1.7\% on ImageNet. Code will be made publicly available.
\end{abstract}

%%%%%%%%% BODY TEXT
\section{Introduction}

\begin{figure}[t]
    \centering
    \small
    \begin{overpic}[width=\linewidth]{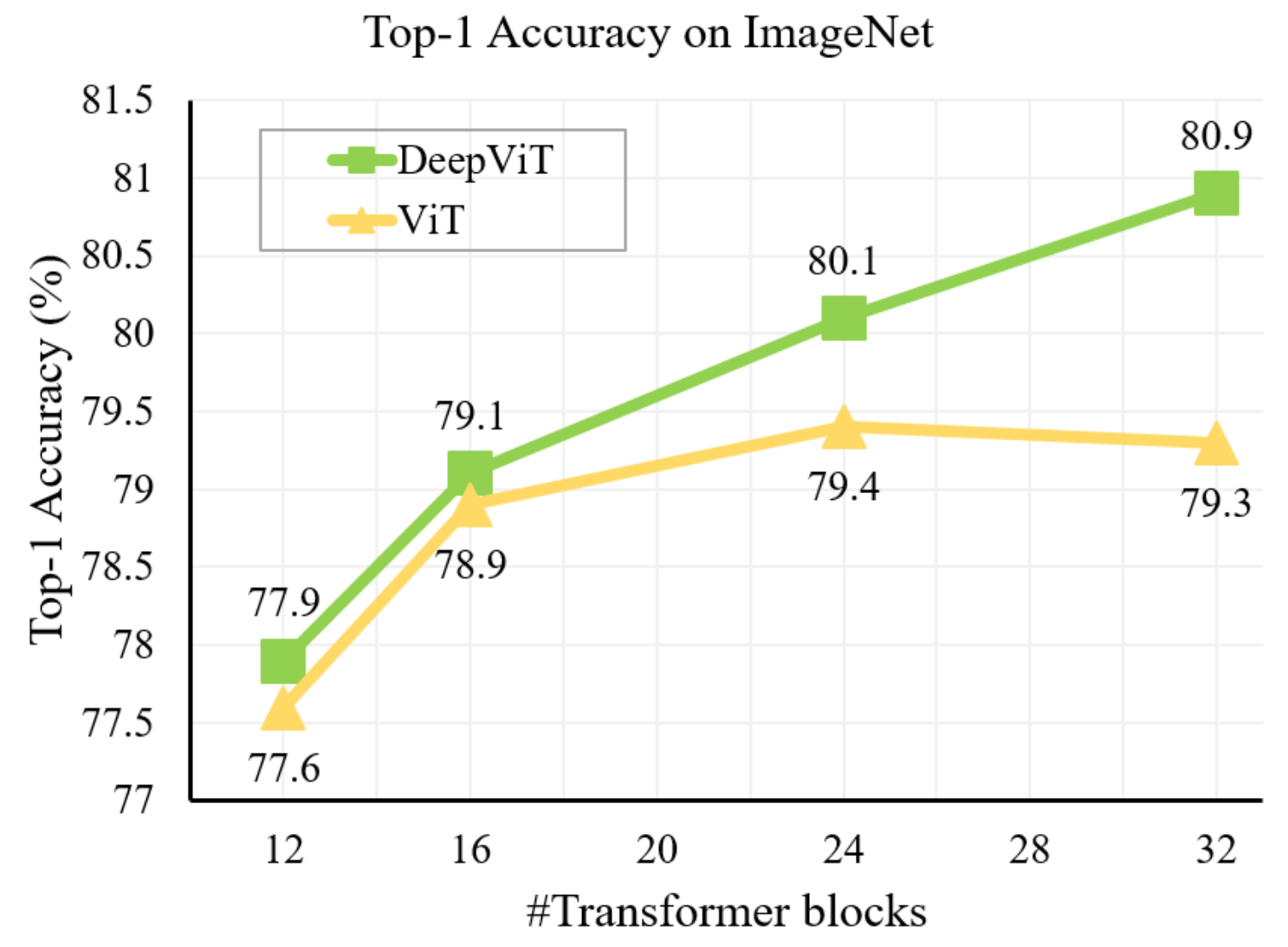}
    \put(36.5, 55.4){\cite{dosovitskiy2020image}}
    \end{overpic}
    \caption{Top-1 classification performance of vision transformers (ViTs) \cite{dosovitskiy2020image} on ImageNet with different network depth 
    \{12, 16, 24, 32\}.
    Directly scaling the depth of ViT by stacking more transformer blocks cannot monotonically  increase the performance. Instead, the model performance  saturates when  going deeper. In contrast, with the proposed \nameofatten, 
    our DeepViT model   successfully achieves better  performance when it goes deeper. 
    }
    \label{fig:depth_vs_acc}
\end{figure}

Recent studies \cite{dosovitskiy2020image,touvron2020training} have demonstrated that transformers
\cite{vaswani2017attention} can be successfully applied to vision tasks \cite{krizhevsky2012imagenet} with competitive
performance compared with convolutional neural networks
(CNNs)~\cite{he2016deep,tan2019efficientnet}.
Different from CNNs that aggregate global information by stacking multiple  convolutions (\eg, $3\times3$)
\cite{he2016deep,he2016identity}, 
vision transformers (ViTs) \cite{dosovitskiy2020image} take  advantages of the self-attention (SA) mechanism \cite{vaswani2017attention} to capture spatial patterns and  non-local dependencies.
This allows ViTs to aggregate rich global information without handcrafting layer-wise local feature extractions as  CNNs and thus achieves better performance. 
%as many layers as  CNNs. \textcolor{red}{[JS: if this is true, why we want to make  ViT deeper?]}
%
For example, as shown in \cite{touvron2020training}, a 12-block ViT model with 22M learnable
parameters achieves better results than the ResNet-101 model which has more than 30 bottleneck
convolutional blocks in ImageNet classification. 
%
% Benefiting from the self-attention mechanism, current

The recent progress of deep CNN models is largely driven by training  very deep models with a large number of layers which is enabled by novel model architecture designs \cite{he2016deep,xie2017aggregated,srinivas2021bottleneck,liu2020improving,zhang2020resnest}. 
% The recent progress of deep CNN models is largely driven by novel model architecture designs, which enable the training of very deep models with a large number of layers  \cite{he2016deep,xie2017aggregated,huang2017densely,zhang2020resnest}.
%
This is because a deeper CNN can learn richer and more complex representations for the input images and provide better performance on vision tasks \cite{bengio2013representation,zhang2018network,rebuffi2017icarl}. Thus, how to effectively scale CNNs to be deeper is an important theme in recent deep learning fields, which stimulates the   techniques like residual learning \cite{he2016deep}. 

Considering the remarkable    performance of shallow ViTs, a natural question arises: 
\emph{can we further improve performance of  ViTs by making it deeper, just like  CNNs?} Though it  seems to be  straightforward at the first glance, the answer may not be trivial since ViT is essentially different from CNNs in its heavy reliance on the self-attention mechanism. 
To settle the question, we  investigate in detail the scalability of ViTs along depth in this work.

% Self-attention based models have been widely used in natural language processing (NLP) tasks \cite{devlin2018bert,vaswani2017attention}. Motivated by the success of those models, recent works have attempted to build purely transformer-based models \cite{touvron2020training,dosovitskiy2020image} without the involvement of convolutional neural networks (CNN). One of the benefits is that self attention models minimize the domain inductive bias \cite{khan2021transformers} and thus is a potential model to unify the architectures for computer vision and natural language processing tasks.

We start with a pilot study on ImageNet to investigate how the performance of ViT changes with increased model depth. In Fig.~\ref{fig:depth_vs_acc}, we show the performance of ViTs \cite{dosovitskiy2020image}
with different block numbers (green line), ranging from 12 to 32.
% \{12, 16, 24, 32\}.
%
As shown, as the number of transformer blocks increases, the model performance does not
improve accordingly. % as expected.
To our surprise, the ViT model with 32 transformer blocks performs even worse than the one with
24 blocks.
This means that directly stacking more transformer blocks as performed in CNNs~\cite{he2016deep}
is inefficient at enhancing ViT models.
% However, the architecture used in those works are typically pre-defined according to the transformer model design as used in NLP. Although important on architecture design \cite{he2016deep,tan2019efficientnet}, the problem of how to scale the depth of a vision transformer is not well understood yet. Although the shortcut connection has been deployed in transformer block, which is considered as a essential component for training a deep neural network \cite{he2016deep}, is it enough to train a deep vision transformer effectively? In this paper, we study this problem in details. We observed that when training on ImageNet dataset \cite{krizhevsky2012imagenet}, the classification accuracy of the vision transformer saturates after the number of blocks is larger than 16. Further increasing the depth do not contribute to the performance gains.
%
We then dig into the cause of this phenomenon. We empirically observed that as the depth of ViTs increases,
the attention maps, used for aggregating the features for each transformer block,  tend to be overly similar after certain layers, which makes the representations stop evolving after certain layers. We name this specific issue as \emph{attention collapse}.
This indicates that as the ViT goes deeper, the self-attention mechanism becomes less effective for generating diverse attentions to capture rich representations. 

% \textcolor{red}{[JS: need to check whether we have experiments to support this argument,  ``the similarity between each token becomes larger". ]}
% We owe this issue to the recursive use of the self-attention modules. Each self-attention layer   aggregates the information from all the tokens into each score in a dynamically generated attention map. As the network goes deeper, since each token keeps receiving the information from all the tokens, the similarity between each token becomes larger. Thus, the generated relationship between different tokens tend to be uniform. This is detailed in Sec. \ref{subsec:oversmoothing}.
%
% As the depth of the network goes deep, the SoftMax operation will smooth the relation among tokens and thus resulting a uniform distribution of the attention map. This is detailed in Sec. \ref{sunsec:softmax_analysis}.
%

To resolve the attention collapse issue and effectively scale the vision transformer to be deeper, 
% we explore several approaches to enhance diversity of the attention maps, including adding regularization   (drop attentions), enhancing sharpness of the attention maps by tuning the temperature parameters during the self-attention computation. Eventually, 
We present a simple yet effective self-attention mechanism, named as \nameofatten{}.
% which regenerates attention maps by exchanging the information from different attention heads
% in a learnable manner.
%
Our \nameofatten{} takes advantage of the multi-head self-attention(MHSA) structure and regenerates attention maps by exchanging the information from different attention heads in a learnable manner.
Experiments show that, Without any extra augmentation and regularization policies, 
simply replacing the MHSA module in ViTs with \nameofatten{} allows us to 
train very deep vision transformers with even 32 transformer blocks with consistent improvements
as shown in Fig.~\ref{fig:depth_vs_acc}.
In addition, we also provide ablation analysis to help better understand of the role
of \nameofatten{} in scaling vision transformers.
% As shown in Figure~\ref{fig:depth_vs_acc}, 
% could improves the top-1 classification accuracy on ImageNet when using 32 blocks. More results are shown
%However, those pure transformer models on vision tasks typically consumes a large memory due to the high dimension of the token embedding. 

To sum up, our contributions are as follows:
\begin{itemize}
    \setlength\itemsep{0em}
    \item We deeply study the behaviour of vision transformers
    and observe that they cannot continuously benefit from stacking more layers as CNNs. We further identify the underlying reasons behind such a counter-intuitive phenomenon and conclude it as \emph{attention collapse} for the first time.
    % \BY{this part should be rewritten. this contribution can be three-fold: 1) conduct the study, 2) make the observation (deeper model gives no benefits), 3) Identify the issue, \ie, attention collapse}
    % find an interesting phenomenon that the attention map of deep transformer blocks tend to be similar and thus results in an inefficiency.

    \item We present \nameofatten{}, a simple yet effective attention mechanism
    that considers information exchange among different attention heads.
    % \QB{To the best of our knowledge, we are the first to successfully train a 32-block ViT
    % on ImageNet-1k from scratch with consistent performance improvement.}
    % of the current transformer blocks which reduce the similarity across different layers significantly.
    
    \item To the best of our knowledge, we are the first to successfully train a 32-block ViT
    on ImageNet-1k from scratch with consistent performance improvement.
    We show that by replacing the self-attention module with our \nameofatten{}, 
    new state-of-the-art results can be achieved on the ImageNet-1k dataset 
    without any pre-training on larger datasets.
\end{itemize}
\section{Related Work}
\subsection{Transformers for Vision Tasks}

Transformers \cite{vaswani2017attention} are initially used for machine translation which replace the recurrence and convolutions entirely with self-attention mechanisms \cite{ramachandran2019stand,hu2019local,zhao2020exploring,ho2019axial,wang2020axial,jiang2020convbert} and achieve outstanding performance. Later,  transformers become the dominant models   for  various  natural language processing (NLP) tasks \cite{brown2020language,radford2019language,devlin2018bert,liu2019roberta}. Motivated by  their success  on the NLP tasks, recent  researchers attempted to combine the self-attention mechanism into CNNs   for computer vision tasks \cite{wang2018non,carion2020end,chen2020pre,sun2019videobert,lu2019vilbert,zheng2020rethinking,zhao2020point}.. Those achievements also stimulate interests of the community in building purely transformer-based models (without convolutions and inductive bias) for vision tasks.  
%However, those models are not efficient on current hardware accelerator due to the special attention patterns. 
The vision transformer (ViT)~\cite{dosovitskiy2020image} is among the first attempt that uses the pure transformer architecture to achieve competitive performance with CNNs on the image classification task. However, due to the large model complexity, ViT needs to be pre-trained on larger-scale  datasets (e.g., JFT300M) for  performing  well on the ImageNet-1k dataset. To solve the data efficiency issue, DeiT \cite{touvron2020training}   deploys knowledge distillation to   train the model with a larger pre-trained teacher model. In this manner, vision transformer can perform well on ImageNet-1k without the need of pre-training on larger dataset. Differently, in this work, we target at a different problem with ViT, \ie, how to effectively scale ViT to be deeper. We propose  a new design for the self-attention mechanism so that it can perform well on vision tasks without the need of extra data, teacher networks,
and the domain specific inductive bias.

\subsection{Depth Scaling  of CNNs   }
% \textcolor{red}{[JS: need to check whether the references listed here are proper and correct. ]}

Increasing the network depth   of a CNN model is deemed to be an effective way to improve the model performance
\cite{simonyan2014very,szegedy2015going,szegedy2016rethinking,he2016identity,huang2017densely}. 
However, very deep CNNs are generally harder to train to perform significantly better   than the shallow ones in the past \cite{glorot2010understanding,zagoruyko2016wide}.  How to effectively scale up the CNNs in depth was a long-standing and challenging problem \cite{huang2018gpipe}. 
The recent progress of CNNs largely benefits from novel architecture design strategies 
that make training deep CNNs more effective~\cite{he2016deep, tan2019efficientnet, howard2019searching, hou2021coordinate, tan2019mixconv,zhou2020rethinking}.
%to be and narrow neural networks (NN) are typically more efficient but hard to train.  With same number of neurons, a deep NN is considered to have larger representation capability than the shallow one. \cite{eldan2016power}. Thus, increasing the depth has been used as a main approach for increasing the performance of an NN. \cite{he2016deep} 
%Following this rule, most of the well performing CNNs are deep with less parameters assigned in each block. \cite{he2016deep, tan2019efficientnet, howard2019searching, hou2021coordinate, tan2019mixconv,zhou2020rethinking} 
Transformer-alike models have modularized architectures and thus can be easily made deeper by
repeating the basic transformer blocks or
%Thus one can scale the model by directly adding more transformer blocks or 
using larger embedding dimensions \cite{brown2020language,lepikhin2020gshard}. However, those straightforward scaling strategies only work well   with larger datasets and stronger augmentation policies \cite{zhong2020random,zhang2017mixup,yun2019cutmix} to alleviate the brought training difficulties.
In this paper, we observed that with the same dataset, 
the performance of vision transformers do saturate as the network depth rises.
We rethink the self-attention mechanism and present a simple but effective approach
to address the difficulties in scaling vision transformers. 
%One of the transformer's advantages is its scability. This is because all blocks shares the same architecture. Thus one can scale the model by directly adding more transformer blocks or using larger embedding dimensions to adapt for larger datasets \cite{brown2020language,lepikhin2020gshard}. However, those models are typically extremely large (over 100B parameters). This indicates a potential redundancy during the scaling. More importantly, those scaling are typically accompanied with larger datasets and stronger augmentation policies. Thus, the problem of scaling with a given target dataset is not well understood on vision tasks. In this paper, we observed that under the same dataset and augmentation policy, the performance of transformer do saturate quickly  after a threshold.

% \subsection{Image classification}

\section{Revisiting Vision Transformer} \label{sec:revisit_vit}

\begin{figure}[t]
    \centering
    \includegraphics[width=\linewidth]{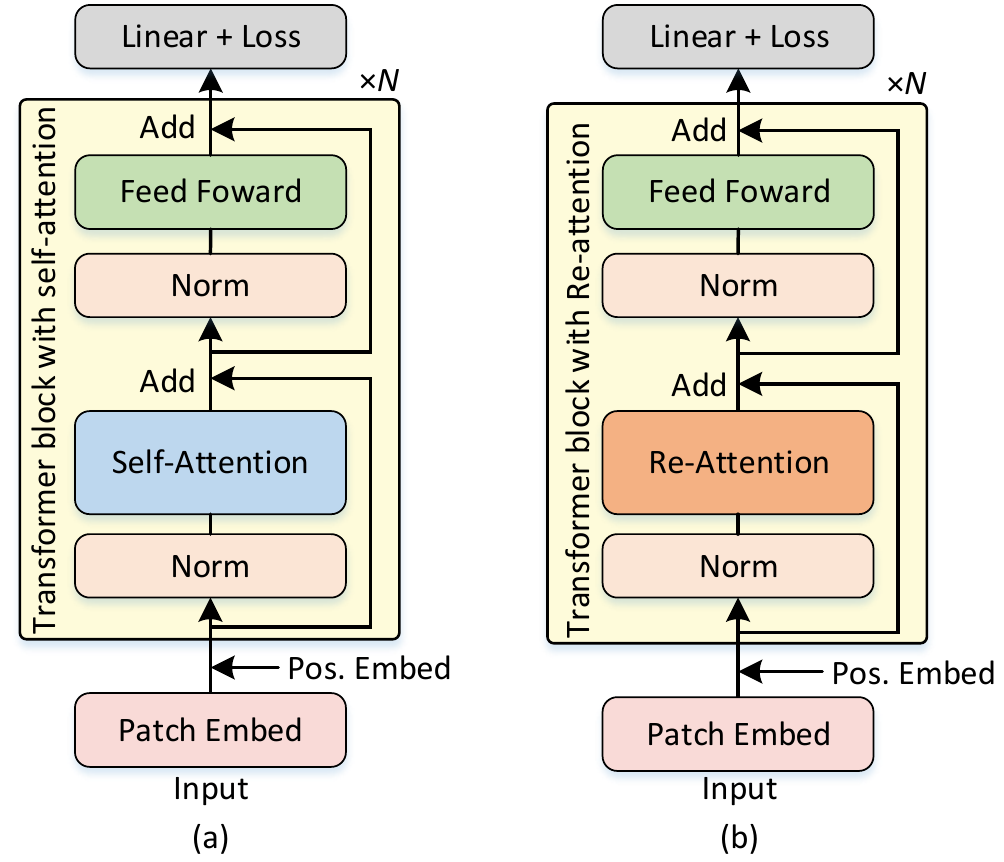}
    \caption{Comparison between the (a) original ViT with $N$ transformer blocks and (b) our proposed DeepViT model. Different from ViT, DeepViT replaces the self-attention layer within  the transformer block  with the proposed \nameofatten{} which effectively addresses the attention collapse issue and enables training deeper ViTs. More details are given in Sec. \ref{subsec:head_gen}.
    }
    \label{fig:diagram}
\end{figure}

A vision transformer (ViT) model \cite{touvron2020training,dosovitskiy2020image}, as depicted in Fig.~\ref{fig:diagram}(a),
is composed of three main components:
a linear layer for patch embedding (\ie, mapping the high-resolution input image
to a low-resolution feature map), a stack of transformer blocks with multi-head self-attention
and feed-forward layers for feature encoding,
and a linear layer for classification score prediction.
% \footnote{As the goal of this paper is to
% explore the scaling capability of ViTs, we keep the original positional embedding part
% unchanged and use the same one as in \cite{dosovitskiy2020image}}. 
In this section, we first review its unique transformer blocks,
in particular the self-attention mechanism, and then we provide studies on 
the collapse problem of self-attention. 
%
% Before explaining the scalable problem of ViT, we first briefly review the transformer block.

% \subsection{Preliminary: Transformer multi-head self attention calculation}
\subsection{Multi-Head Self-Attention}
% \textcolor{red}{[JS: we need a figure to better explain the basic components of ViT. Since ViT is a new model, we need more words to introduce ViT for giving enough background to the readers to understand our work. ]}
% Recent works use transformer for image classification tasks with minimum modifications \cite{dosovitskiy2020image}. 
%
% It comprises three parts: patch embeddings that encode the input image, a 
% The transformer block includes two key components: a self-attention (SA) layer 
% and a multi-layer perceptron (MLP). The self-attention layer is based on
% the mapping of a query ($Q$) and a pair of key ($K$)-value ($V$) pairs to an output,
% where $Q,K,V$ are all calculated based the same set of input features.
%
Transformers \cite{vaswani2017attention} were extensively used in natural language  
for encoding a sequence of input word tokens into a sequence of embeddings. 
To comply with such sequence-to-sequence  learning structure when processing images, 
ViTs first divide an input image into multiple  patches uniformly and
encode each patch into a token embedding. 
Then, all these tokens, together with a class token,  are fed into a stack of transformer blocks. 

Each transformer block consists of a multi-head self-attention (MHSA) layer 
and a feed-forward  multi-layer perceptron (MLP). The MHSA   generates a trainable associate memory with a query ($Q$) and a pair of key ($K$)-value ($V$) pairs to an output
via linearly transforming the input.  
Mathematically, the output of a MHSA is calculated by:
\begin{equation} \label{eqn:self_atten_definition}
% \begin{split}
\text{Attention}(Q,K,V) = \text{Softmax}({QK^\top}/{\sqrt{d}})V,
% \end{split}
\end{equation}
where $\sqrt{d}$ is a scaling factor based on the depth of the network. The output of the MHSA is then normalized and   fed into the MLP to generate the input to the next block. 
%Figure~\ref{fig:diagram}(a) gives a description on the transformer block.
%
In the above self-attention, $Q$ and $K$ are multiplied to generate the attention map, which represents the correlation
between all the tokens within each layer. It is used to retrieve and combine the embeddings in the value  $V$. In the following, we particularly analyze the role of the attention map in   scaling the ViT. For convenience,  we use $\mathbf{A} \in \mathbb{R}^{H \times T \times  T}$ to denote the attention map, with $H$ being the number of SA heads and $T$ the number of tokens. For the $h$-th SA head, the attention map is computed as  $\mathbf{A}_{h,:,:} = \text{Softmax}({Q_h K_{h}} ^\top/\sqrt{d})$ with $Q_h$ and $K_h$ from the corresponding head.
When the context is clear, we omit the subscript $h$.
% \BY{Should we subscript Q and K with head index $h$? and mention that when the context is clear, $h$ might be omitted} 
%

\begin{figure*}[t]
\begin{center}
\includegraphics[width=0.95\linewidth]{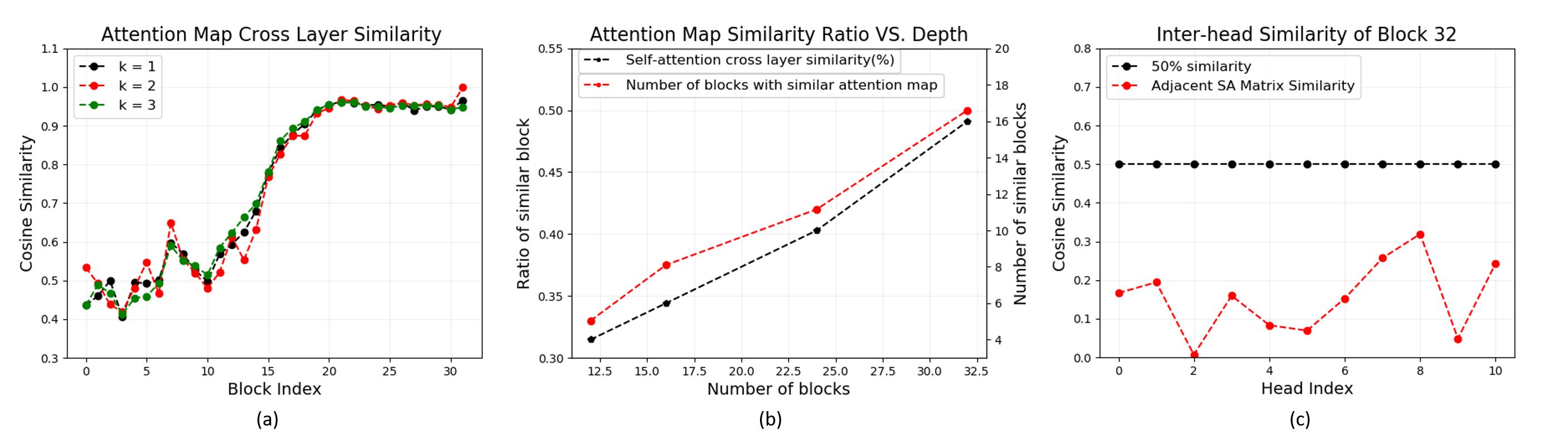}
\vspace{-20pt}
\end{center}
  \caption{(a) The similarity ratio of the generated self-attention maps across different layers. The visualization is based on ViT models with 32 blocks pre-trained on ImageNet. For visualization purpose, we plot the ratio of token-wise attention vectors with similarity in Eqn.~\eqref{eqn:cross_layer_similarity} larger than   the average similarity within nearest $k$  transformer blocks. As can be seen, the similarity ratio is   larger than 90\% for blocks after the 17th one. (b) The ratio  of similar blocks to the total number of blocks increases when the depth of the ViT model increases.   (c) Similarity of attention maps from different heads within the same block. The similarity between different heads within the blocks is all lower than 30\% and they present sufficient diversity.}
  \vspace{-10pt}
\label{fig:atten_similarity}
\end{figure*}

\subsection{Attention Collapse}

% \textcolor{red}{[JS: The first paragraph is redundant with the introduction. Here in the first paragraph,  highlight CNNs benefit from increasing model depth. We conduct a systematic study to see how ViT performs by simply increasing its depth.]} 
% Different from CNNs that exploit convolutions with normally $3\times3$ kernel size
% to encode spatial information, vision transformers often leverage self-attention to model
% long-range dependencies.
% %
% A strong advantage of adopting self-attention is that only a dozen of self-attention
% layers are needed when building vision transformers \cite{touvron2020training}
% to achieve competitive performance to powerful CNNs \cite{he2016deep}.
% %
% Regarding the scaling capability of CNNs \cite{he2016deep,he2016identity,tan2019efficientnet},
% could we empower the vision transformers via scaling in the depth dimension?
% %
% We thus first study the   effectiveness of scaling vision transformers
% in depth.

Motivated by the success of deep CNNs \cite{he2016deep,simonyan2014very, tan2019efficientnet, tan2019mixconv}, we conduct   systematic study in the changes of the performance of ViTs as depth increases.
% We first study the scaling capability of the vision transformer that is proposed in \cite{dosovitskiy2020image}. 
Without loss of generality, we first fix the hidden dimension  and 
the number of heads to 384 and 12 respectively\footnote{Similar phenomenon can also
be found when we vary the hidden dimension size according to our experiments.}, following the common  practice in~\cite{touvron2020training}.
% To speed up the training process, we use smaller embedding dimension of the base model and keep all the other architecture and training settings the same.
% We also adopt the improvements as proposed in timm library \cite{rwightman}. 
%
Then we stack different number of transformer blocks (varying from 12 to 32) to build multiple ViT models corresponding to different depths. The overall performances for image classification are evaluated on ImageNet~\cite{krizhevsky2012imagenet} and summarized in Fig.~\ref{fig:depth_vs_acc}.
As evidenced by the performance curve, we surprisingly find that the classification accuracy improves slowly and saturates fast as the model goes deeper. More specifically, we can observe that the improvement stops after employing 24 transformer blocks. This phenomenon demonstrates that  existing ViTs have difficulty in gaining benefits from deeper architectures. 
% Surprisingly, we find the classification accuracy saturates fast. When the model has 24 transformer blocks, the performance is not improved anymore.
%

% To figure out the root reason of such a scaling problem of ViTs,
%we first focus on one of the major components of the transformer block
Such a problem is quite counter-intuitive and worth exploration, as similar issues (\ie, how to effectively train a deeper  model) have also been observed for CNNs at its early development stage~\cite{he2016deep}, but properly solved later \cite{he2016deep,he2016identity}. By taking a deeper look into the transfromer architecture, we would like to highlight that the self-attention mechanism plays a key role in ViTs, which makes it significantly different from CNNs. Therefore,  we  start with  investigating how the self-attention, or more concretely, the generated attention map $\mathbf{A}$ varies as the model goes deeper. 
%To this end, we propose to first study how the self-attention matrix varies as the network goes deeper.
%
%the self-attention module from two orthogonal perspectives. 1) Whether and how is the attention learned from different head at the same layer different with each other. 2) Whether and how is the attention learned of the same head from different layers different with each other.
% \textcolor{red}{[JS: need some motivation here on why we choose to look into the attention maps.]}

To measure the evolution of the attention maps over layers, we compute the following
cross-layer similarity between the attention maps from different layers:
% \begin{equation}
% % \label{eqn:cross_layer_similarity}
% %     M^{l,q}_{i,j} = \frac{\mathbf{A}^l_{:,i,:} \cdot \mathbf{A}^{q}_{:,i,:}}{|\mathbf{A}^l_{:,i,:}|  |\mathbf{A}^{q}_{:,j,:}|},
% % \end{equation}
% \label{eqn:cross_layer_similarity}
%     M^{p,q}_{i,j} = \sum_{t=1}^{T} \frac{ {\mathbf{A}^p_{:,i,t}}^{\top}  \mathbf{A}^{q}_{:,j,t}}{\|\mathbf{A}^p_{:,i,t}\|  \|\mathbf{A}^{q}_{:,j,t}\|},
% \end{equation}
% where $M^{p,q}$ is the cosine similarity matrix of the self-attention map between the layers $p$ and $q$ and each element $M^{p,q}_{i,j}$ is the similarity of self-attention for tokens $i$ and $j$ with all the other tokens across all the heads.
\begin{equation}
\label{eqn:cross_layer_similarity}
    M^{p,q}_{h,t} = \frac{ {\mathbf{A}^p_{h,:,t}}^{\top}  \mathbf{A}^{q}_{h,:,t}}{\|\mathbf{A}^p_{h,:,t}\|  \|\mathbf{A}^{q}_{h,:,t}\|},
\end{equation}
where $M^{p,q}$ is the cosine similarity matrix between the attention map of layers $p$ and $q$. Each element $M^{p,q}_{h,t}$ measures the similarity of attention for head $h$ and token $t$. {Consider one specific self-attention layer and its $h$-th head, $\mathbf{A}^{*}_{h,:,t}$ is a $T$-dimensional vector representing how much the input token $t$ contributes to each of the $T$ output tokens. 
%For example, $\mathbf{A}^{*}_{h,t,k} = 0$ means the input token $t$ is not considered at all when the output token $k$ is obtained. 
$M^{p,q}_{h,t}$, thus, provides an appropriate measurement on how the contribution of one token varies from layer $p$ to $q$. When $M^{p,q}_{h,t}$ equals one, it means that token $t$ plays exactly the same role for self-attention in layers $p$ and $q$.}

% To measure the similarity between the self-attention maps, we define the similarity $S^{l,q}$ between the self-attention maps of two layers as the ratio of the number of similar vector pairs to the total number of pairs between two self-attention maps:
% \begin{equation}
%     S(l,q) = \frac{\sum I}{|M^{l,q}|}, \quad I_{i,j} =
%     \begin{cases}
%       1, & \text{\quad if $M^{l,q}_{i,j}$ $>$ $\tau$}\\
%       0, & \text{\quad otherwise}
%     \end{cases}  
% \end{equation}
% %
% Where $\tau$ is a hyper-parameter and used as a threshold for deciding similar vectors\footnote{0.5 is selected as a threshold for visualization purpose in this paper}. 

Given Eqn.~\eqref{eqn:cross_layer_similarity}, we then train a ViT model with 32 transformer blocks on ImageNet-1k and  investigate the above similarity  among all the attention maps. As shown in Fig.~\ref{fig:atten_similarity}(a), the ratio of similar attention maps in  $M$ after the 17th block is larger than 90\% .
%\textcolor{red}{[JS: this is misleading. Not similarity larger than 90\% but the ratio of similar attention maps.]}. 
This indicates that the learned attention maps afterwards are similar and the transformer block may degenerate to an MLP. As a result, further stacking such degenerated MHSA
may introduce the model rank degeneration issue (\ie, the rank of  the model parameter tensor  from multiplying the layer-wise parameters together will decrease) and limits the model learning capacity. This is also validated by our analysis on the degeneration of  learned features as shown below.
%MLPs would
%adding more SA modules at deep layers may not contribute new information useful for improving the model performance.  
Such observed attention collapse could be one of the reasons for the observed performance saturation of ViTs. 
% If this is the case, this phenomenon should be severer as the network goes deeper. 
% \textcolor{red}{[the transition here is not smooth]} 
% This indicates an inefficiency when scaling along the depth dimension. 
% \textcolor{red}{[the transition here is not smooth]} 
% However, how does this phenomenon change with the depth of the network? 
To further validate the existence of  this phenomenon for ViTs with different depths, we  conduct the same experiments on ViTs with 12, 16, 24 and 32 transformer blocks respectively and calculate the number of blocks with similar attention maps. The results  shown in Fig.~\ref{fig:atten_similarity}(b) clearly demonstrate the  ratio of the number of similar attention map blocks to the total number of blocks increases when adding more transformer blocks. 

\begin{figure}[t] 
\small
\begin{center}
\includegraphics[width=0.49\linewidth]{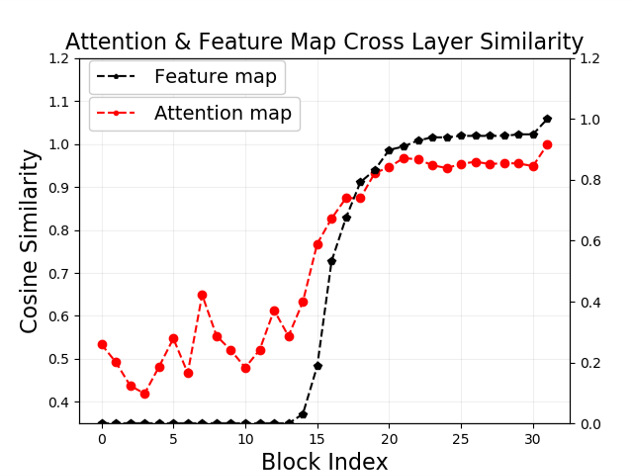}
\includegraphics[width=0.49\linewidth]{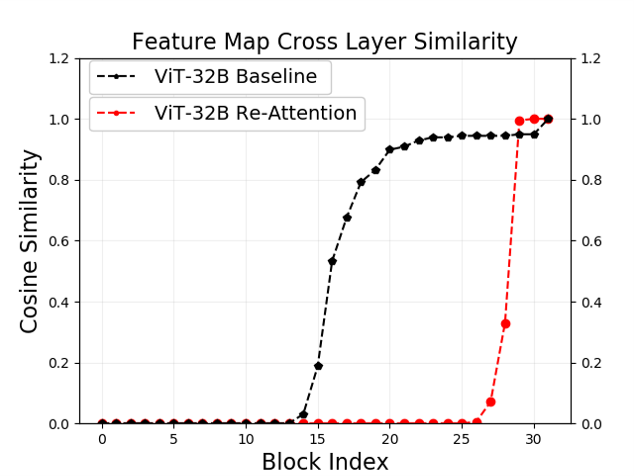}
\end{center}
\vspace{-15pt}
  \caption{
  (\textbf{Left}): Cross layer similarity of attention map and features for ViTs. The black dotted line shows the cosine similarity between feature maps of the last block and each of the previous blocks. The red dotted line shows the ratio of similar attention maps of adjacent blocks. The visualization is based on a 32-block ViT model   pre-trained on ImageNet-1k. (\textbf{Right}): Feature map cross layer cosine similarity for both the original ViT model and ours. As can be seen, replacing
  the original self-attention with \nameofatten{} could reduce the feature map similarity significantly. % Cosine similarity of the final features output from the 32nd block with the output features from  all the previous blocks for a 32-block ViT model   trained on ImageNet-1k. % \textcolor{red}{[JS: the caption of the plot is not correct. Remove the y-axis label on the right side. Change attention matrix to attention map in the legend for terminology consistency.  ]}
  }
\label{fig:feature_similarity}
% \vspace{-3mm}
\end{figure}

%\BY{May consider to add the feature similarity visualization here.}
To understand how the  attention collapse may hurt the ViT model performance, we further  study how it affects feature learning of the deeper layers. 
%more direct effects of the utilization of self-attention, \ie, how the representation learned is actually affected. To this end, 
{For a specific 32-block ViT model, we compare the final output features  with the outputs of  each intermediate transformer block  %\textcolor{red}{[JS: need to be more specific. the features are output from MLP? ]} 
by investigating their cosine similarity. The results in Fig.~\ref{fig:feature_similarity}  demonstrate that the similarity is quite high and the  learned features stop evolving 
after the 20th block. There is a close correlation between the increase of attention similarity and feature similarity.} 
%in Fig.~\ref{fig:atten_similarity}.
This observation indicates that attention collapse is responsible for the non-scalable issue of ViTs.

\section{\nameofatten{} for Deep ViT}

As revealed above, one major obstacle in scaling up ViT to a deeper one is the attention
collapse problem. In this section, we present two solution approaches, one is to increase the hidden dimension for computing self-attention and the other one is a novel re-attention mechanism. 

\subsection{Self-Attention in Higher Dimension Space}
One intuitive solution to conquer attention collapse is to increase the embedding dimension of each token. This will augment the representation capability of each token embedding to encode more information. As such, the resultant attention maps can be more diverse and the similarity between each block's attention map could be reduced. {Without loss of generality, we verify this approach empirically by conducting a set of experiments based on ViT models with 12 blocks for quick experiments.} 
%\textcolor{red}{[JS: need to discuss why we take 12-block for this study, instead of 32 (most of previous observations are made based on 32-block model.]}
%  with varying token embedding dimensions. 
Following previous transformer based works \cite{vaswani2017attention,dosovitskiy2020image}, four embedding dimensions are selected,  ranging from 256 to 768.  The detailed configurations and the results are shown in Tab.~\ref{tab:increasing_dim}. 

\begin{figure}[t] 
\begin{center}
\includegraphics[width=0.8\linewidth]{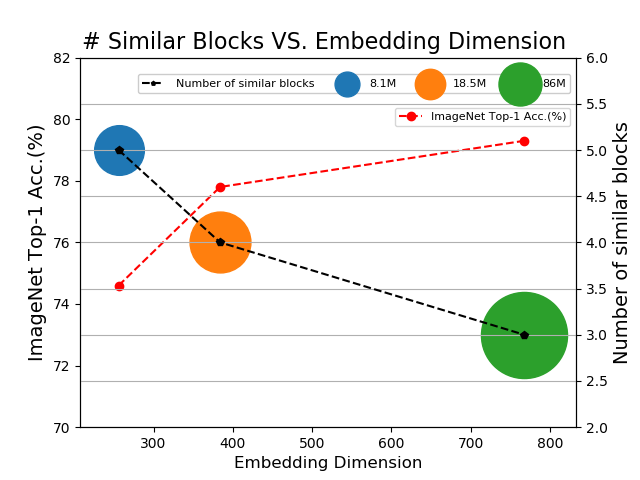}
\end{center}
\vspace{-15pt}
  \caption{Impacts of embedding dimension on the similarity of generated self-attention map across layers. As can be seen, the number of similar attention maps decreases with increasing embedding dimension. However, the model size also increases rapidly. 
  %\textcolor{red}{[JS: caption needs to update. embedding number means embedding dimension? give embedding dimension in the figure. also plot the corresponding model performance. ]}
  }
\label{fig:similar_block_vs_embedding}
% \vspace{-2mm}
\end{figure}

\begin{table}[t]
\footnotesize
\caption{Top-1 accuracy on ImageNet-1k dataset of vision transformer with different embedding dimensions. The number of model parameters increase quadratically with the embedding dimension. The number of similar attention map blocks with different embedding dimensions are shown in Figure \ref{fig:similar_block_vs_embedding}}. 
\label{tab:increasing_dim}
% \begin{center}
\centering
\setlength\tabcolsep{1.7mm}
\renewcommand\arraystretch{1.0}
\begin{tabular}{ccccc}
\toprule
% \multicolumn{1}{c}{\bf Group}  
\bf Model
% & \bf Searching Algo.
&\bf \#Blocks
&\bf Embed Dim.
&\bf \#Param. (M)
&\bf Top-1 Acc.(\%)
\\ \midrule %\\
% \multirow{10}{*}{\makecell[l]{w/o SE}} 
% & MBV2-0.75 &- & 2.6 & 209 & 69.8 \\
 ViT  & 12 & 256 & 8.15 & 74.6 \\
 ViT  & 12 & 384 & 18.51 & 77.86 \\
 ViT  & 12 & 512 & 33.04 & 78.8 \\
 ViT  & 12 & 768 & 86 & 79.3 \\
% &Ours$^*$(Voted)& Gradient + Manual &  5.22 & 420 & 77.88\\
\bottomrule
\vspace{-5mm}
\end{tabular}
% \end{center}
\end{table}

% \begin{figure*}[t]
% \begin{center}
% % \fbox{\rule{0pt}{2in} \rule{0.9\linewidth}{0pt}}
% \includegraphics[width=0.99\linewidth]{figures/block_diagram.png}
% \end{center}
%   \caption{Our proposed head re-generation process.  }
% \label{fig:inter_head_similarity}
% \end{figure*}

From Fig.~\ref{fig:similar_block_vs_embedding} and Tab.~\ref{tab:increasing_dim}, one can see that the number of blocks with similar attention maps is  reduced and the attention collapse is alleviated by increasing the  embedding dimension. Consequently, 
the model performance is also increased accordingly. This validates our core hypothesis\textemdash the attention collapse is the main bottleneck for scaling ViT. Despite its effectiveness, increasing the embedding dimension also increases the computation cost   significantly and the brought performance  improvement tends to diminish. Besides, a larger model (with higher embedding dimension) typically needs  more data for training, suffering  the   over-fitting risk  and   decreased  efficiency. 

% In Sec. \ref{exp:experiments}, we show that with the same classification accuracy, a deeper model with smaller embedding could be 2$\times$ smaller than a ViT model with larger embedding dimension. 
% In this work, we aim to propose a method that could reduce the similarity of attention maps  with as less computational   overhead. This motivates us to develop the \nameofatten{} from another perspective, as detailed below.
%More details are shown in Sec. \ref{subsec:head_gen}.

\subsection{\nameofatten{}}
\label{subsec:head_gen}

It has been demonstrated in Sec.~\ref{sec:revisit_vit} that the similarity between
attention maps across different transformer blocks is high, especially for deep
layers.
However, we find  the similarity of attention maps from different heads of
the same transformer block is quite small, as shown in Fig.~\ref{fig:atten_similarity}(c).
%
%In Figure~\ref{fig:atten_similarity}(c), we depict the similarity among different heads of the last transformer block from a 32-block ViT.
%
% We develop the \nameofatten{} based on our following empirical observations. Interestingly, we observed that although the similarity between attention maps across different blocks  are high, the similarity of attention maps from different  heads within the same block is small, as shown in Figure~\ref{fig:inter_head_similarity}.
Clearly, different heads from the same self-attention layer 
focus on different aspects of the input tokens.
Based on this observation, we propose to establish cross-head communication 
to re-generate the attention maps and train deep ViTs to perform better. 
% We measure the cosine similarity between adjacent heads and the results are shown in Figure \ref{fig:inter_head_similarity}. 

% \begin{figure}[t]
% \begin{center}
% % \fbox{\rule{0pt}{2in} \rule{0.9\linewidth}{0pt}}
% \includegraphics[width=0.8\linewidth]{figures/inter_head_similarity_v2.png}
% \end{center}
% \vspace{-15pt}
%   \caption{Inter-head similarity of generated self-attention map of block 32. As shown in Figure \ref{fig:atten_similarity}, the attention matrix similarity between block 32 and other blocks are larger thab 90\%. However, the similarity between different heads within the blocks is all lower than 30\%. }
% \label{fig:inter_head_similarity}
% \end{figure}

Concretely, we  use the attention maps from the heads as basis and generate a new set of attention maps by dynamically aggregating them. To achieve this, we define a learnable transformation matrix $\Theta \in \mathbb{R}^{H \times H}$ and then use it to mix the multi-head attention maps into  re-generated new ones,  before being multiplied with $V$. Specifically, the \nameofatten{} is implemented by:
\begin{equation}
\label{eqn:head_regen}
    \text{Re-Attention}(Q,K,V) = \text{Norm}(\Theta^\top (\text{Softmax}(\frac{QK^\top}{\sqrt{d}})))V, 
\end{equation}
where transformation matrix $\Theta$ is multiplied to the self-attention map $\mathbf{A}$ along the head dimension. Here $\mathrm{Norm}$ is a normalization
function used to reduced the layer-wise variance. $\Theta$ is end-to-end learnable. 
%

% \QB{We'd better give a few lines of code about re-attention.}

\myPara{Advantages:} 
The advantages of the proposed \nameofatten{} are two-fold.
First of all, compared with other  possible attention augmentation methods,
such as randomly dropping some elements of the attention map or tuning  SoftMax temperature,
our \nameofatten{} exploits the interactions among different
attention heads  to collect their complementary information  and better improves the attention map diversity. This is  also verified by our  following experiments.
Furthermore, our \nameofatten{} is effective and easy to implement.
It needs only a few lines of code and negligible computational overhead
compared to the original self-attention. Thus it is much more efficient than the approach of increasing embedding dimension. 
%
% In our experiment section, we will give more analysis.
% Two show the advantages of \nameofatten{}, we set two commonly used methods for injecting variations to the attention matrix: dropout and SoftMax temperature. More details and definitions on the implementation are shown in Sec. \ref{subsubsec:temperature}. We run extensive experiments and verify that this simple modifications could reduce the cross layer self-attention map similarity and thus improve the performance significantly as the model goes deep, as shown in Sec.~\ref{exp:add_variations}. 

% However, shallower layers prefers Softmax for local feature extraction

% \subsection{Effects of Softmax}
% \label{sunsec:softmax_analysis}

% \subsection{Dual self-attention block}
% \subsection{Local-Global attention block}

\section{Experiments}
\label{exp:experiments}

% In this section, we conduct a series of experiments to answer the following questions: (a) is our proposed \nameofatten~ effective on training deep vision transformers? (b)   does our proposed method also work well for  shallow ViT models? \textcolor{red}{[JS: the question (c) comes without context thus cannot understand why it is important.   ]}(c) is our proposed method compatible with CNN patch embeddings as the one proposed in \cite{yuan2021tokens}?  To answer these questions, we first conduct experiments to verify the   highly similar attention maps indeed contain redundant information. Then we present multiple methods to resolve this issue and evaluate them via comprehensive experiments. Finally, we compare the proposed deep ViT model against the lastest state-of-the-arts

In this section, we first conduct experiments to further demonstrate the attention collapse
problem. Then, we give extensive ablation analysis to show the advantages of the proposed \nameofatten{}. By incorporating \nameofatten{} into the transformers, we design two modified version of vision transformers and name them as deep vision transformers (DeepViT). Finally, we compare the proposed DeepViT models against the latest state-of-the-arts (SOTA).

\subsection{Experiment Details}

% \textcolor{red}{[JS: re-arrange the order of these tables. some tables are mentioned earlier in the text but are put behind others.]}

% \myPara{Dataset and training hyper-parameters:}

To make a fair comparison, we first tuned a set of parameters 
for training the ViT base model and then use the same set of hyper-parameters for all the ablation experiments.
Specifically, we use AdamW optimizer \cite{loshchilov2017decoupled} and cosine  learning rate decay policy with an
initial learning rate of 0.0005. We use 8 Telsa-V100 GPUs and train the model for 300 epochs using Pytorch \cite{paszke2019pytorch} library. 
The batch size is set to 256.  
% We apply exponential moving average when training the network to speed up the convergence. 
We use 3 epochs for learning rate warm-up \cite{loshchilov2016sgdr}. We also use some  augmentation techniques such as mixup \cite{zhang2017mixup} and random augmentation \cite{cubuk2020randaugment} to boost the performance of baseline models following \cite{zhang2020resnest}. 
% Those settings are advised in timm library \cite{rwightman} for reproducing ViT models. 
%
When comparing with other methods, we adopt the same set of hyper-parameters as used by the target models. 
We report results on the ImageNet dataset \cite{krizhevsky2012imagenet}. For all experiments, the image size is set to be 224$\times$224. 
%\QB{List the hyper-parameters one by one. Tell when using T2T to achieve SOTA.}
%
% \myPara{Baseline ViT models:}
To study the scaling capability of current transformer blocks,
% we choose two ViT models with 24, 32 and 48 \textcolor{red}{[JS: we will have 48?]} transformer blocks respectively. To speed up the training   and reduce the GPU memory costs, 
we set the embedding dimension to 384 and the expansion ratio 3
for the MLP layers. We use 12 heads for all the models. More detailed configurations are shown in Tab. \ref{tab:ablation_depth}.

\begin{table}[h]
\footnotesize
\caption{Baseline model specifications. All ablation experiments are based on the ViT models with different number of blocks. The `\#B' in `ViT-\#B' denotes the number of transformer blocks in the model.}
\label{tab:ablation_depth}
% \begin{center}
\centering
\begin{tabular}{lcccc}
\toprule
% \multicolumn{1}{c}{\bf Group}  
\bf Model
% & \bf Searching Algo.
&\bf \#Blocks
&\bf \#Embeddings

&\bf MLP Size
&\bf Params. (M)
\\ \midrule %\\
% \multirow{10}{*}{\makecell[l]{w/o SE}} 
% & MBV2-0.75 &- & 2.6 & 209 & 69.8 \\
 ViT-16B & 16 & 384 & 1152 & 24.46 \\
 ViT-24B & 24 & 384 & 1152 & 36.26 \\
 ViT-32B & 32 & 384 & 1152 & 48.09 \\
% &Ours$^*$(Voted)& Gradient + Manual &  5.22 & 420 & 77.88\\
\bottomrule
\vspace{-5mm}
\end{tabular}
% \end{center}
\end{table}

\subsection{More Analysis on Attention Collapse}

% In Sec.~\ref{sec:revisit_vit}, we have pointed out the collapse problem of self-attention
% by showing the cross-layer similarity of self-attention maps and the ratio of similar
% blocks for different depths of ViTs (Fig.~\ref{fig:atten_similarity}(a-b)).
In this section, we show more  analysis on the attention map similarity  and study how the collapsed attention maps affect the model performance. 
% support this. \textcolor{red}{[JS: ``to support this'' is not concrete. we need say more specifically why we carry out the following experiments.]}

\myPara{Attention reuse:}
As discussed above, when the model goes deeper, the attention maps of the deeper blocks 
become highly similar. This implies that adding more blocks on a deep ViT model may not improve  the model performance.
To further verify this claim, we design an experiment to reuse the attention maps computed 
at an early block of ViT to replace the ones after it. 
% Namely, we   let those similar attention maps  share the same set of values.
%
Specifically, we run experiments on the ViT models with 24 blocks and 32 blocks
but share the $Q$ and $K$  values (and the resulted attention maps) of the  last ``unique'' block   to  all the blocks afterwards.
The ``unique'' block is defined as the block whose  attention map's similarity ratio
with adjacent layers is smaller than 90\%. % \QB{Make sure this is correct.}
More implementation details can be found in the supplementary material. 
The results are shown in Tab.~\ref{tab:ablation_sharing}. 
Surprisingly, for a ViT model with 32 transformer blocks, when we use the same  $Q$ and $K$ values for 
%directly share the $Q$ and $K$ values of 
the last 15 blocks, the performance degradation is negligible.
This implies the attention collapse problem indeed exists
and reveals the inefficacy in adding more blocks when the model is deep.  

\begin{table}[h]
\footnotesize
\caption{ImageNet top-1 accuracy of the ViT models with shared self-attention maps. `\#Shared blocks' denotes the number of the transformer blocks that share the same attention map.  }
\label{tab:ablation_sharing}
% \begin{center}
\centering
\begin{tabular}{ccccc}
\toprule
% \multicolumn{1}{c}{\bf Group}  
\bf \#Blocks
% & \bf Searching Algo.
&\bf \#Embeddings
% &\bf \# Heads
&\bf \#Shared blocks
&\bf Top-1 Acc. (\%)
\\ \midrule %\\
% 12 & 768 & 0 & 79.3 \\
% 12 &  768 & 4 & 79.5 \\ 
% \midrule
24 & 384 & 0 & 79.3 \\
24 &  384 & 11 & 78.7 \\ 
\midrule
32 &  384   & 0 & 79.2 \\ 
32 &  384   & 15 & 79.2 \\ 
% &Ours$^*$(Voted)& Gradient + Manual &  5.22 & 420 & 77.88\\
\bottomrule
\vspace{-5mm}
\end{tabular}
% \end{center}
\end{table}

\myPara{Visualization:} %\QB{Maybe need more analysis.}
To more intuitively  understand   the attention map collapse across layers, we visualize the learned attention maps from different blocks of the original ViT \cite{dosovitskiy2020image}.
We take a 32-block ViT model as an example and pre-train it on ImageNet. The visualization of the attention maps with original MHSA and \nameofatten{}  are shown 
in Fig.~\ref{fig:vis_atten_map}. 
% The first row shows the attention map of original ViT model and the second row shows the results with \nameofatten{}.
%
% More visualizations can be found in our supplementary material.
%
% \textcolor{cyan}
{It can be observed that the original MHSA learns the local relationship  among the adjacent 
  patches in  the shallow blocks and the attention maps tend to expand to cover more patches gradually. In the deep blocks, the MHSA learns   nearly uniform global attention maps with high similarity. Differently, after implementing \nameofatten{}, the attention maps at deep blocks keep the diversity and have small similarities from adjacent blocks.}

\renewcommand{\addFig}[1]{\includegraphics[width=0.135\linewidth]{figures/atten_visualize/#1}}
\renewcommand{\addFigs}[1]{\addFig{layer_#1_head_mean_bs_mean.jpg}}
\newcommand{\addFigsRe}[1]{\addFig{layer_#1_head_mean_bs_mean_re.jpg}}

\begin{figure*}[t]
  \centering
  \footnotesize
  \setlength\tabcolsep{0.2mm}
  \renewcommand\arraystretch{1.0}
  \begin{tabular}{cccccccc}
    \rotatebox{90}{~Self-attention} & 
    \addFigs{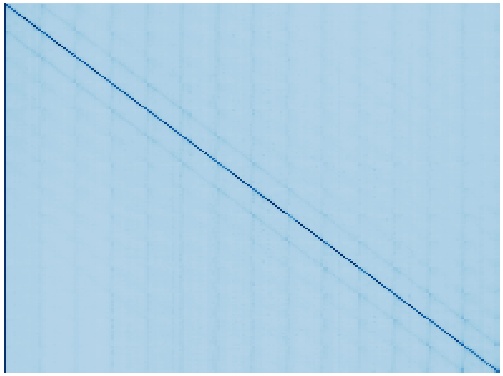} & 
    \addFigs{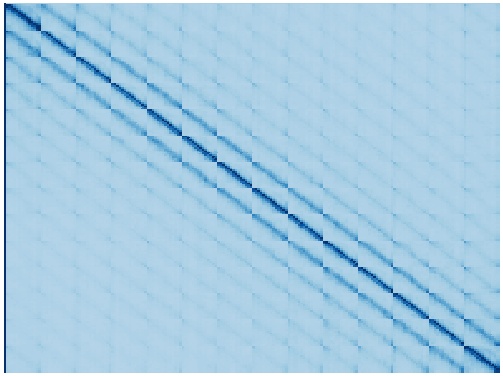} & 
    \addFigs{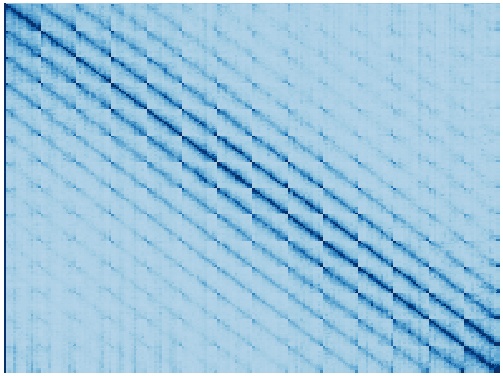} & 
    \addFigs{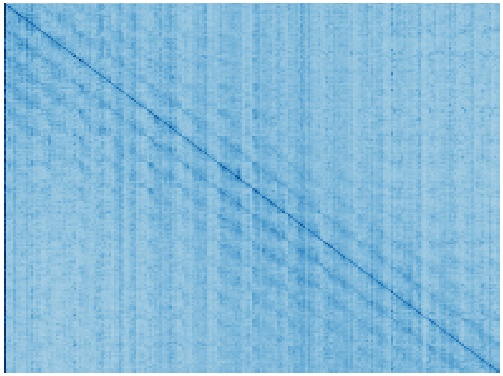} & 
    \addFigs{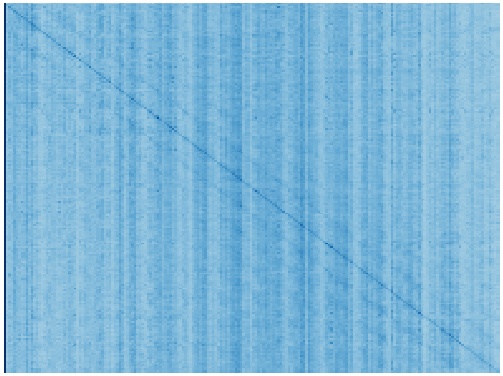} &
    \addFigs{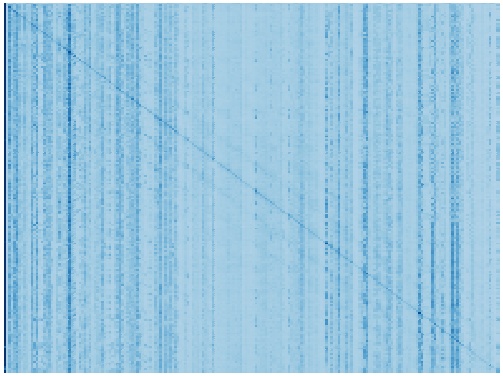} &
    % \addFigs{5} & 
    % \addFigs{6} & 
    % \addFigs{7} & 
    % \addFigs{8} &
    \addFigs{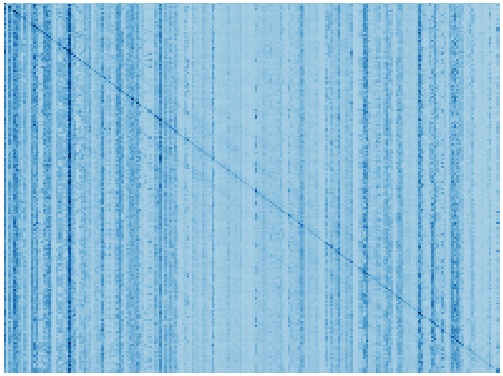} \\
    \rotatebox{90}{~~Re-attention} & 
    \addFigsRe{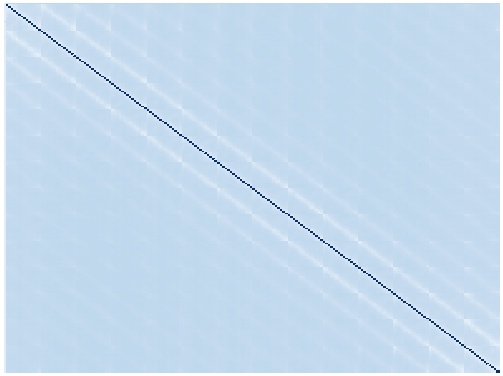} & 
    \addFigsRe{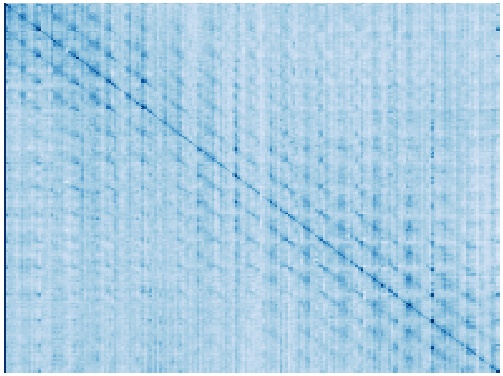} &
    \addFigsRe{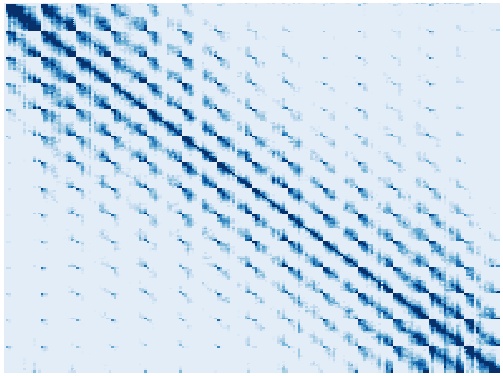} & 
    \addFigsRe{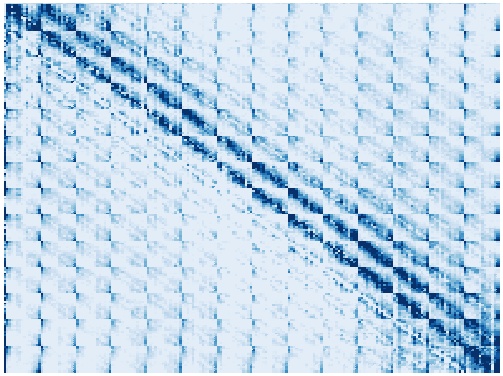} & 
    \addFigsRe{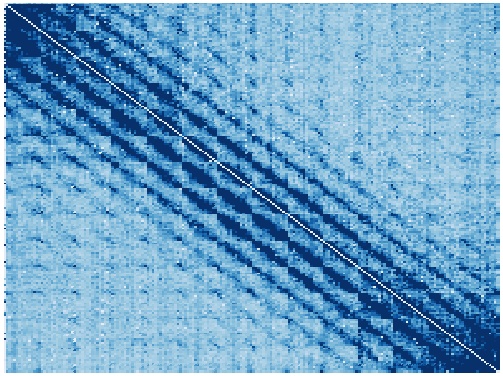} &
    \addFigsRe{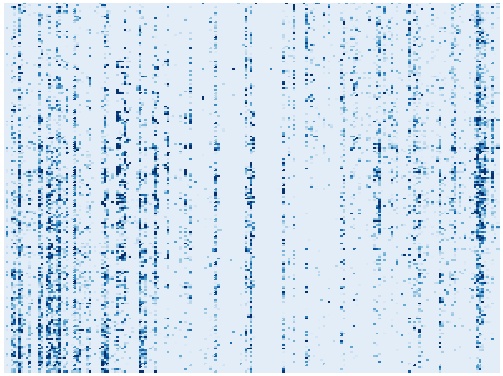} &
    % \addFigs{15} & 
    % \addFigs{16} & 
    % \addFigs{17} & 
    % \addFigs{18} &
    \addFigsRe{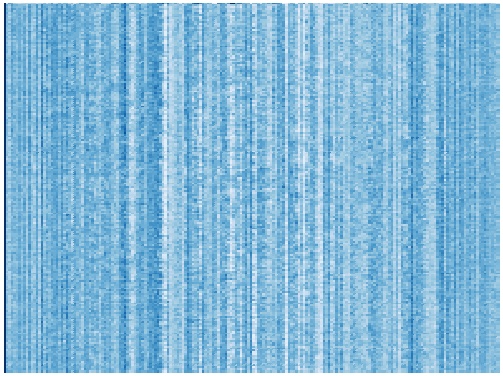} \\
     & Block 1 & Block 4 & Block 11 & Block 18  &Block 23 & Block 29 
     & Block 30
     \\
  \end{tabular}
  \vspace{3pt}
  \caption{Attention map visualization of the selected blocks of the baseline  ViT model with 32 transformer blocks. The first row is based on original Self-attention module and the second is based on \nameofatten{}. As can be seen, the model   only learns local patch relationship at its  shallow blocks with the rest of attention values  near to zero. Though their the scope  increases gradually as the block goes deeper,  the attention maps tend to become   nearly uniform and thus lose diversity. After adding \nameofatten{}, the originally similar attention maps are changed to be diverse as shown in the second row. Only at the last block's attention map, a nearly uniform attention map is learned. %\textcolor{red}{[JS: add discussion on the attention maps of \nameofatten{}.]}
  }
  \label{fig:vis_atten_map}
  %\vspace{-10pt}
\end{figure*}

% \myPara{Adjacent SA similarity Analysis:}

\subsection{Analysis on \nameofatten{}}
\label{exp:add_variations}

{In this subsection, we present two   straightforward  modifications to the current self-attention mechanism  as baselines.  We then conduct a series of comparison experiments to show the advantages
of our proposed \nameofatten{}.}
% in generating more diverse attention maps by comparing it with two straightforward methods: (a) adding temperature to the SoftMax layer and (b) applying dropout on the self-attention maps.%  as detailed in Sec. \ref{subsubsec:temperature} and Sec. \ref{subsubsec:dropatten} respectively. 

\begin{figure}[t]
    \centering
    \includegraphics[width  =\linewidth]{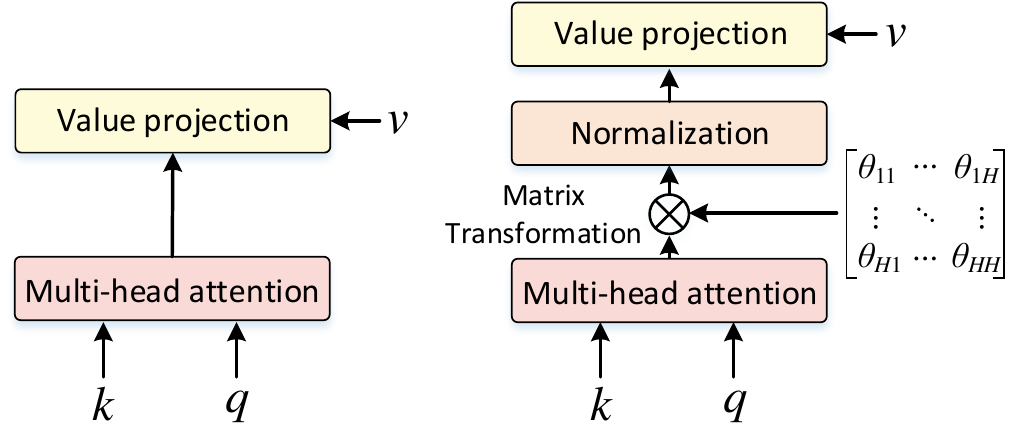}
    \caption{(\textbf{Left}): The original self-attention mechanism; (\textbf{Right}): Our proposed re-attention mechanism. As shown, the original attention map is mixed via a learnable matrix $\Theta$ before multiplied with values.}
    \label{fig              :reatt}
\end{figure}

\myPara{Re-attention v.s. Self-attention:}
% \subsubsection{Deep ViT models with \nameofatten{}}
We first evaluate the effectiveness of \nameofatten{} by comparing to the pure ViT models
using the same set of training hyper-parameters.
We directly replace the self-attention module in ViT with \nameofatten{} and show the results
in Tab.~\ref{tab:head_regen} with different number of transformer blocks.
As can be seen, the vanilla ViT architecture suffers   performance saturation
when adding more transformer blocks. 
This phenomenon coincides with our observations that the number of blocks with similar attention maps increases with the depth as shown in Fig. \ref{fig:atten_similarity}(b). 
% It is also observed that the number of
%  blocks with similar attention maps 
% increases with the depth as well.
%
Interestingly, when replacing the   self-attention with
our proposed \nameofatten{}, the number of similar blocks are all reduced to be zero and the performance rises consistently
as the model depth increases.
The performance gain is especially significant for deep ViT with 32 blocks. This might be explained by the fact that the 32 block ViT model has the largest number of blocks with similar attention maps and the improvements should be proportional to the number similar blocks in the model.
%
% In addition, the number of similar blocks decreases to 0 after using the proposed \nameofatten{}.
%
% In Figure~\ref{fig:vis_atten_map} (bottom), we visualize
% the attention maps produced by our \nameofatten{}.
% %
% The similarity among different blocks is low even for
% deep blocks.
%
These experiments demonstrate that the proposed \nameofatten{}
can indeed solve the attention collapse problem
and thus enables training a very deep vision transformer without extra datasets or augmentation policies.

\begin{table}[h]
\footnotesize
\caption{ImageNet Top-1 accuracy of deep ViT (DeepViT) models with \nameofatten{} and different number of transformer blocks. }
\label{tab:head_regen}
% \begin{center}
\centering
\begin{tabular}{lcccc}
\toprule
% \multicolumn{1}{c}{\bf Group}  
\bf Model
% & \bf Searching Algo.
&\bf \#Similar Blocks
&\bf Param. (M)
&\bf Top-1 Acc. (\%)
\\ \midrule %\\
% \multirow{10}{*}{\makecell[l]{w/o SE}} 
% & MBV2-0.75 &- & 2.6 & 209 & 69.8 \\
 ViT-16B \cite{dosovitskiy2020image} & 5 & 24.5 & 78.9 \\
 DeepViT-16B & 0  & 24.5 & 79.1 (+0.2) \\ 
\midrule
 ViT-24B \cite{dosovitskiy2020image} & 11 & 36.3 & 79.4 \\
 DeepViT-24B & 0  & 36.3 & 80.1 (+0.7) \\ 
 \midrule
 ViT-32B \cite{dosovitskiy2020image} & 16 & 48.1 & 79.3 \\
 DeepViT-32B & 0  & 48.1 & 80.9 (+1.6) \\ 
% &Ours$^*$(Voted)& Gradient + Manual &  5.22 & 420 & 77.88\\
\bottomrule
\vspace{-5mm}
\end{tabular}
% \end{center}
\end{table}

% As the self-attention is the output of SoftMax layers, all the deep layers thus have similar attentions map and this  reduces  the contribution to the model performance. To verify this, we   visualize  the self-attention maps of the ViT-32l model for the first 30 blocks in Figure~\ref{fig:vis_atten_map}. Clearly,   after the 19$^{th}$ block, the attention maps shows a nearly uniform distribution and all of them look quite  similar. 

% \myPara{Network depth}
% Add analysis. Both shallow and deep models work.

\begin{figure*}[h]
\begin{center}
\includegraphics[width=1.\linewidth]{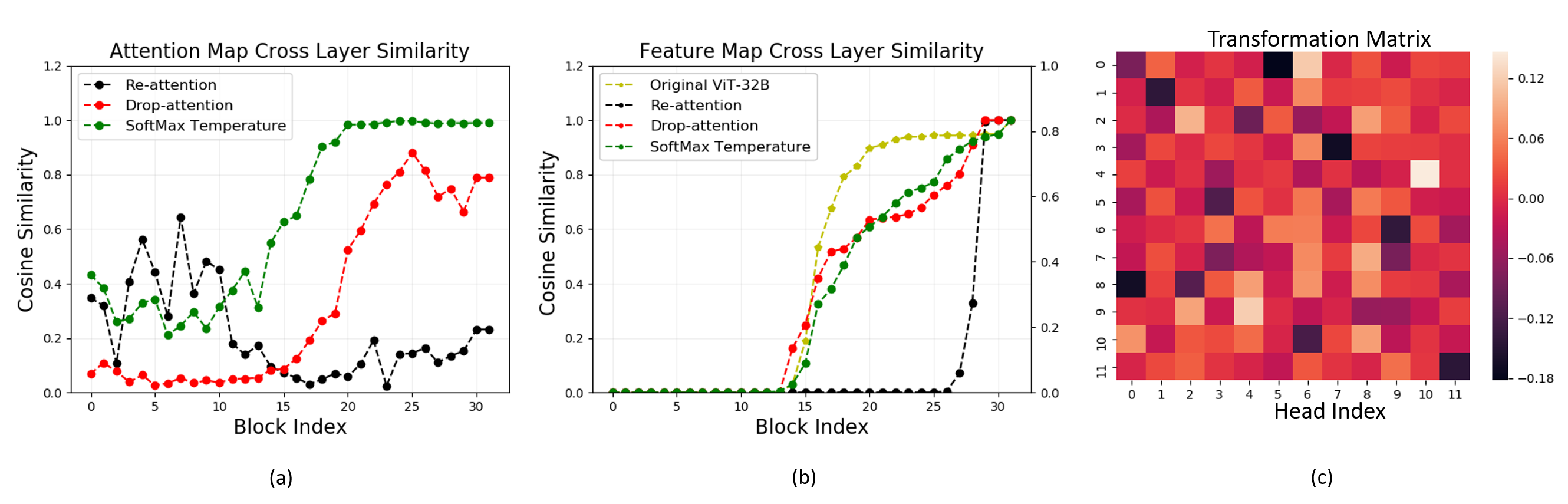}
\end{center}
\vspace{-20pt}
  \caption{(a) Adjacent block attention map similarity with different methods. As can be seen, our proposed \nameofatten{} achieves low cross layer attention map similarity. 
  (b) Cosine similarity between the feature map of the last block and each of the previous block. 
  (c) Visualization of transformation matrix of the last block.}
\label{fig:fc_weights_heatmap}
\end{figure*}

\myPara{Comparison to adding temperature in self-attention:}
The most intuitive way to mitigate the over-smoothing phenomenon is to sharpen  the distribution of the elements in the attention map of MHSA. We could achieve this by assigning a   temperature $\tau$  to the Softmax layer of MHSA:
\begin{equation} \label{eqn:self_atten_definition_temp}
% \begin{split}
\text{Attention}(Q,K,V) = \text{Softmax}\left(\frac{QK^\top}{\tau \sqrt{d}}\right)V.
% \end{split}
\end{equation}
As the attention collapse is observed to be severe on deep layers (as shown in Fig.~\ref{fig:atten_similarity}),
%\textcolor{red}{[JS: point to the relevant results]}
 we design two sets of experiments on a ViT model with 32 transformer blocks: (a)   linearly decaying the temperature $\tau $   in each block such that the attention map distribution is sharpened and (b) making the  temperature learnable and optimized together with the model training. We first check the impact of the SoftMax temperature on reducing the attention map similarity. As shown in Fig. \ref{fig:fc_weights_heatmap}(a), the number of similar blocks are still large. Correspondingly, the feature similarity among blocks are also large as shown in Fig. \ref{fig:fc_weights_heatmap}(b). Thus, adding a temperature to the SoftMax only reduces the attention map similarity by a small margin.   The classification results on ImageNet are shown in Tab.~\ref{tab:ablation_temperature}. As shown, using a learnable temperature could improve the performance but the improvement is marginal.

\begin{table}[h]
\footnotesize
\caption{ImageNet Top-1 accuracy of the ViT models with  SoftMax temperature $\tau$ and the drop attention. The embedding dimension of all the models is set as 384. }
\label{tab:ablation_temperature}
% \begin{center}
\centering
\begin{tabular}{lcccc}
\toprule
% \multicolumn{1}{c}{\bf Group}  
\bf \# Blocks
% & \bf Searching Algo.
&\bf \# Similar Blocks
&\bf Model
% &\bf \# Shared Blocks
&\bf Acc. (\%)
\\ \midrule %\\
% \multirow{10}{*}{\makecell[l]{w/o SE}} 
% & MBV2-0.75 &- & 2.6 & 209 & 69.8 \\
%  24 & 384 & 12 & 0 & 79.88 \\
% 24 &  384 & 12 & 12 & 78.7 \\ 
% \midrule
32 &  16  & Vanilla & 79.3 \\ 
32 &  13  & Linearly decayed $\tau$ & 79.0  \\ 
32 &  10  & Learnable $\tau$  & 79.5  \\ 
32 &  8  & drop attention & 79.5  \\ 
32 & 0   & \nameofatten{} & 80.9 \\
% &Ours$^*$(Voted)& Gradient + Manual &  5.22 & 420 & 77.88\\
\bottomrule
\vspace{-5mm}
\end{tabular}
% \end{center}
\end{table}

\paragraph{Comparison to dropping attentions:}
% \label{subsubsec:dropatten}
Another baseline we have attempted to differentiate the self-attention maps across layers is to use random dropout on the attention maps $\mathbf{A}$. 
As the dropout will mask out    different   positions on the attention maps for different blocks, the similarity between attention maps could be reduced.
% As for different blocks  dropout will mask out    different   positions on the attention maps. This will  reduce the similarity between attention maps.  
The impacts on the attention maps and the output features of each block are shown in Fig. \ref{fig:fc_weights_heatmap}(a-b). It is observed that dropping attention does reduce the cross layer similarity of the attention maps. However, the similarity among features are not reduced by much. This is because the difference between attention maps comes from the zero positions in the generated mask. Those zero values do reduce the similarity between attention maps but not contribute to the features. Thus, the improvement is still not significant as shown in Tab.~\ref{tab:ablation_temperature}.
% The results are given in Tab.~\ref{tab:ablation_temperature}. Again the improvement is still not significant.  

% \begin{table}[h]
% \footnotesize
% \caption{ViT model with dropout attention.}
% \label{tab:ablation_temperature}
% % \begin{center}
% \centering
% \begin{tabular}{lcccc}
% \toprule
% % \multicolumn{1}{c}{\bf Group}  
% \bf \# Blocks
% % & \bf Searching Algo.
% &\bf \# Embedding
% &\bf Drop Attention
% % &\bf \# Shared Blocks
% &\bf Acc.(\%)
% \\ \midrule %\\
% % \multirow{10}{*}{\makecell[l]{w/o SE}} 
% % & MBV2-0.75 &- & 2.6 & 209 & 69.8 \\
% %  24 & 384 & 12 & 0 & 79.88 \\
% % 24 &  384 & 12 & 12 & 78.7 \\ 
% % \midrule
% 32 &  384 & \ding{55} & 79.25 \\ 
% 32 &  384 & \checkmark & 79.46  \\ 
% % &Ours$^*$(Voted)& Gradient + Manual &  5.22 & 420 & 77.88\\
% \bottomrule
% \vspace{-5mm}
% \end{tabular}
% % \end{center}
% \label{tab:drop_atten}
% \end{table}

% \myPara{Analysis on the transformation matrix $\Theta$:}  The key component of \nameofatten{} is the transformation matrix $\Theta$. It aggregates self-attention information  of   different  heads   and generates   new attention maps. This simple modification incorporate the inter-head communication that is not covered by the original MHSA design. It is worth noting that the original MHSA deisgn can be thought as a special case of \nameofatten{} with a diagonal transformation matrix. By making $\Theta$ learnable for each block, an optimized pattern could be learned end to end during the training process.

% \myPara{Visualization}

\myPara{Advantages of \nameofatten{}:}
Our proposed \nameofatten{} brings  more significant improvements over the temperature-tuning and the attention-dropping  methods. This is because both adding temperature and dropping  attention are regularizing the distribution of the originally over-smoothed self-attention maps, without explicitly encouraging them to be diverse. However, our proposed \nameofatten{} mechanism
 uses different heads (whose attention maps are dissimilar) as basis and  re-generate the attention maps via the transformation matrix $\Theta$. This process incorporates the inter-head information communication and the generated attention maps can encode richer information. It is worth noting that the original MHSA design can be thought as a special case of \nameofatten{} with an identity  transformation matrix. By making $\Theta$ learnable for each block, an optimized pattern could be learned end to end.  As shown in Fig.~\ref{fig:fc_weights_heatmap}(c), the learned transformation matrix assigns a diverse set of weights for each newly generated head. It clearly shows that the combination for each new heads takes different weights from the original heads in the re-attention process and thus reduces the   similarity between their attention maps. As shown in Fig.~\ref{fig:fc_weights_heatmap}(a), our proposed \nameofatten{} achieves the lowest cross layer attention map similarity. Consequently, it also reduces the feature map similarity across layers as shown in Fig.~\ref{fig:fc_weights_heatmap}(b).

\subsection{Comparison with other SOTA models}
% \paragraph{Comparison with ViT} To verify the effectiveness of \nameofatten{}, we directly apply \nameofatten{} to ViT-base \cite{dosovitskiy2020image} model. With the same set of training hyper-parameters, adding \nameofatten{} could improve the performance by 0.62\% as shown in Table \ref{tab:vit_base_apply_transform}. 

% \begin{table}[t]
% \footnotesize
% \caption{Direct comparison with ViT-base model. We directly replace the self-attention with \nameofatten{}. The two sets of experiments are trained using the same set of training recipes.}
% \label{tab:vit_base_apply_transform}
% % \begin{center}
% \centering
% \begin{tabular}{lcccc}
% \toprule
% % \multicolumn{1}{c}{\bf Group}  
% \bf Model
% % & \bf Searching Algo.
% &\bf Params. (M)
% &\bf MAdds (M)
% &\bf Acc. (\%)
% \\ \midrule %\\
% % \multirow{10}{*}{\makecell[l]{w/o SE}} 
% % & MBV2-0.75 &- & 2.6 & 209 & 69.8 \\
%  ViT-Base & 86M & 18G & 79.3 \\
% \midrule
% ViT-Base + ours & 86M  & 18G & 79.92\\ 
% % &Ours$^*$(Voted)& Gradient + Manual &  5.22 & 420 & 77.88\\
% \bottomrule
% % \vspace{-5mm}
% \end{tabular}
% % \end{center}
% \end{table}

% Since the attention collapse mainly comes from the deep blocks, we could only apply \nameofatten{} on the blocks that has similar attention maps. We have conducted   analysis on where to apply \nameofatten{} to ViT models and the related  details and experiment results are given in the supplementary material. 

% a potential separation point
With \nameofatten{}, we design two ViT variants, \ie,  DeepViT-S and DeepViT-L,  based on the ViT with 16 and 32 transformer blocks respectively. For both   models, we use \nameofatten{} to replace the   self-attention. To have a similar number of parameters with other ViT models, we adjust the embedding dimension accordingly. The hidden dimensions of  DeepViT-S and DeepViT-L models are set as 396 and 408 respectively. More details on the model configuration are given in the supplementary material.
 Besides, motivated by \cite{yuan2021tokens}, we add   three CNN layers for extracting  the token embeddings, using the same configurations as     \cite{yuan2021tokens}.  It is worth noting that we do not use the optimized training recipes and the repeated augmentation as  \cite{touvron2020training} for training our models.  The results are shown in Tab.~\ref{tab:sota_comparison}.  Clearly, our DeepViT model achieves higher accuracy with less  parameters than the recent CNN  and ViT based models. Notably, without any complicated architecture change as made by T2T-ViT \cite{yuan2021tokens} (adopting a deep-narrow architecture)  or  DeiT \cite{touvron2020training} (introducing token distillation), simply using the \nameofatten{} makes our DeepViT-L outperforms them  by 0.4 points  with even smaller model size (55M vs.\ 64M \& 86 M).

% \begin{table*}[t]
% \footnotesize
% \caption{Architecture of DeepVit small and large}
% \label{tab:sota_comparison_arch}
% % \begin{center}
% \centering
% \begin{tabular}{lccccccc}
% \toprule
% % \multicolumn{1}{c}{\bf Group}  
% \bf Model
% % & \bf Searching Algo.
% &\bf \# Blocks
% &\bf \# Embedding
% &\bf \# Heads
% &\bf Split ratio
% &\bf Param.(M)
% &\bf MAdds (M)
% &\bf Acc. (Top-1 \%)
% \\ \midrule %\\
% % \multirow{10}{*}{\makecell[l]{w/o SE}} 
% % & MBV2-0.75 &- & 2.6 & 209 & 69.8 \\
%  DeepVit-small & 16 & 384 & 12 & 11-5 & 27.3M & & 82.3 \\
% DeepVit-large &  32 & & 12 &  &  &  &\\ 
% % &Ours$^*$(Voted)& Gradient + Manual &  5.22 & 420 & 77.88\\
% \bottomrule
% \vspace{-5mm}
% \end{tabular}
% % \end{center}
% \end{table*}

\begin{table}[t]
\footnotesize
\caption{Top-1 accuracy comparison with other SOTA models on ImageNet. * denotes our reproduced results. $^{\star}$ denotes our model trained with training recipes used in DeiT \cite{touvron2020training}. }
\label{tab:sota_comparison}
% \begin{center}
\centering
\begin{tabular}{lcccc}
\toprule
% \multicolumn{1}{c}{\bf Group}  
\bf Model
% & \bf Searching Algo.
&\bf Params. (M)
&\bf MAdds (G)
&\bf Acc. (\%)
\\ \midrule %\\
% \multirow{10}{*}{\makecell[l]{w/o SE}} 
% & MBV2-0.75 &- & 2.6 & 209 & 69.8 \\
 ResNet50 \cite{he2016deep} & 25 & 4.0 & 76.2 \\
 ResNet50* & 25 & 4.0 & 79.0 \\
 RegNetY-8GF \cite{radosavovic2020designing} & 40 & 8.0 & 79.3 \\
 Vit-B/16 \cite{dosovitskiy2020image} & 86 & 17.7 & 77.9 \\
 Vit-B/16* & 86 & 17.7 & 79.3 \\
 T2T-ViT-16 \cite{yuan2021tokens} & 21 & 4.8 & 80.6 \\
 DeiT-S \cite{touvron2020training} & 22 & -- & 79.8 \\
%  DeiT-S (KD) \cite{touvron2020training} & 22 & -- & 81.2 \\
 \midrule
%  DeepVit-S (Ours) & 27  &  & 82.3 \\ 
  DeepVit-S (Ours) & 27  & 6.2 & 81.4 \\ 
  DeepVit-S$^{\star}$ (Ours) & 27  & 6.2 & 82.3 \\ 
\midrule
 ResNet152 \cite{he2016deep} & 60 & 11.6 & 78.3 \\
 ResNet152* & 60 & 11.6 & 80.6 \\
 RegNetY-16GF \cite{radosavovic2020designing} & 54 & 15.9 & 80.0 \\
 Vit-L/16 \cite{dosovitskiy2020image} & 307 & -- & 76.5 \\
 T2T-ViT-24 \cite{yuan2021tokens} & 64 & 12.6 & 81.8 \\
 DeiT-B \cite{touvron2020training} & 86 & -- & 81.8 \\
 DeiT-B* & 86 & 17.7 & 81.5 \\
%  DeiT-B (KD) \cite{touvron2020training} & 86 & -- & 83.4 \\
\midrule
%  DeepVit-L (Ours) & 55  &  & \\ 
 DeepVit-L (Ours) & 55  & 12.5 & 82.2 \\ 
 DeepVit-L$^{\star}$ (Ours) & 58  & 12.8 & 83.1 \\ 
 DeepVit-L$^{\star} \uparrow$ 384 (Ours) & 58  & 12.8 & 84.3 \\ 
% &Ours$^*$(Voted)& Gradient + Manual &  5.22 & 420 & 77.88\\
\bottomrule
\vspace{-5mm}
\end{tabular}
% \end{center}
\end{table}

% \paragraph{Comparison with DeiT} As shown in \cite{touvron2020training}, a finetuned traiing recipe could improve the performance of ViTs significantly. To verify the efficiency of our method, we apply the same set of training parameters to re-train the model. The results are shown in Table \ref{tab:comaprison_deit}.

% \begin{table}[t]
% \footnotesize
% \caption{Direct comparison with DeiT models. \textcolor{red}{[JS: why do we need this individual table? ]}}
% \label{tab:comaprison_deit}
% % \begin{center}
% \centering
% \begin{tabular}{lcccc}
% \toprule
% % \multicolumn{1}{c}{\bf Group}  
% \bf Model
% % & \bf Searching Algo.
% &\bf Params. (M)
% &\bf MAdds (M)
% &\bf Acc. (\%)
% \\ \midrule %\\
%  DeiT-S \cite{touvron2020training} & 22 &  & 79.8 \\
%  DeiT-S (KD) \cite{touvron2020training} & 22 &  & 81.2 \\
%  \midrule
%  DeepVit-S (Ours) & 27  &  & 82.3 \\ 
% \bottomrule
% \vspace{-5mm}
% \end{tabular}
% % \end{center}
% \end{table}

% \begin{figure}[t]
% \begin{center}
%   %\includegraphics[width=0.8\linewidth]{egfigure.eps}
% \end{center}
%   \caption{Example of caption.  It is set in Roman so that mathematics
%   (always set in Roman: $B \sin A = A \sin B$) may be included without an
%   ugly clash.}
% \label{fig:long}
% \label{fig:onecol}
% \end{figure}

\section{Conclusion}

% In this work, we observe an interest phenomenon that increasing the depth of the current vision transformers cannot increase the performance with the same set of training recipes. 
% We conduct extensive experiments to identify  that one of the causes of the in-efficacy on depth-scaling comes from the attention collapse\textemdash the attention maps become  highly similar  at the deep blocks. 
In this work, we found the attention collapse problem of vision transformers as they go deeper and
propose a novel \nameofatten{} mechanism to solve it with minimum amount of computation and memory overhead. 
With our proposed \nameofatten, we are able to maintain an increasing performance when increasing the depth of ViTs. 
%Transformer has achieved on par performance with CNNs on vision tasks. However, the architecture of the transformer block has not been well explored. 
% Based on our results, we believe more research on understanding the rationale of transformer blocks will bring further gains.
We hope our observations and methods could facilitate the development of vision transformers
in future.

{\small
\bibliographystyle{ieee_fullname}
\bibliography{egbib}

\begin{thebibliography}{10}\itemsep=-1pt

\bibitem{bengio2013representation}
Yoshua Bengio, Aaron Courville, and Pascal Vincent.
\newblock Representation learning: A review and new perspectives.
\newblock {\em IEEE transactions on pattern analysis and machine intelligence},
  35(8):1798--1828, 2013.

\bibitem{brown2020language}
Tom~B Brown, Benjamin Mann, Nick Ryder, Melanie Subbiah, Jared Kaplan, Prafulla
  Dhariwal, Arvind Neelakantan, Pranav Shyam, Girish Sastry, Amanda Askell,
  et~al.
\newblock Language models are few-shot learners.
\newblock {\em arXiv preprint arXiv:2005.14165}, 2020.

\bibitem{carion2020end}
Nicolas Carion, Francisco Massa, Gabriel Synnaeve, Nicolas Usunier, Alexander
  Kirillov, and Sergey Zagoruyko.
\newblock End-to-end object detection with transformers.
\newblock In {\em European Conference on Computer Vision}, pages 213--229.
  Springer, 2020.

\bibitem{chen2020pre}
Hanting Chen, Yunhe Wang, Tianyu Guo, Chang Xu, Yiping Deng, Zhenhua Liu, Siwei
  Ma, Chunjing Xu, Chao Xu, and Wen Gao.
\newblock Pre-trained image processing transformer.
\newblock {\em arXiv preprint arXiv:2012.00364}, 2020.

\bibitem{cubuk2020randaugment}
Ekin~D Cubuk, Barret Zoph, Jonathon Shlens, and Quoc~V Le.
\newblock Randaugment: Practical automated data augmentation with a reduced
  search space.
\newblock In {\em Proceedings of the IEEE/CVF Conference on Computer Vision and
  Pattern Recognition Workshops}, pages 702--703, 2020.

\bibitem{devlin2018bert}
Jacob Devlin, Ming-Wei Chang, Kenton Lee, and Kristina Toutanova.
\newblock Bert: Pre-training of deep bidirectional transformers for language
  understanding.
\newblock {\em arXiv preprint arXiv:1810.04805}, 2018.

\bibitem{dosovitskiy2020image}
Alexey Dosovitskiy, Lucas Beyer, Alexander Kolesnikov, Dirk Weissenborn,
  Xiaohua Zhai, Thomas Unterthiner, Mostafa Dehghani, Matthias Minderer, Georg
  Heigold, Sylvain Gelly, et~al.
\newblock An image is worth 16x16 words: Transformers for image recognition at
  scale.
\newblock {\em arXiv preprint arXiv:2010.11929}, 2020.

\bibitem{glorot2010understanding}
Xavier Glorot and Yoshua Bengio.
\newblock Understanding the difficulty of training deep feedforward neural
  networks.
\newblock In {\em Proceedings of the thirteenth international conference on
  artificial intelligence and statistics}, pages 249--256. JMLR Workshop and
  Conference Proceedings, 2010.

\bibitem{he2016deep}
Kaiming He, Xiangyu Zhang, Shaoqing Ren, and Jian Sun.
\newblock Deep residual learning for image recognition.
\newblock In {\em Proceedings of the IEEE conference on computer vision and
  pattern recognition}, pages 770--778, 2016.

\bibitem{he2016identity}
Kaiming He, Xiangyu Zhang, Shaoqing Ren, and Jian Sun.
\newblock Identity mappings in deep residual networks.
\newblock In {\em European conference on computer vision}, pages 630--645.
  Springer, 2016.

\bibitem{ho2019axial}
Jonathan Ho, Nal Kalchbrenner, Dirk Weissenborn, and Tim Salimans.
\newblock Axial attention in multidimensional transformers.
\newblock {\em arXiv preprint arXiv:1912.12180}, 2019.

\bibitem{hou2021coordinate}
Qibin Hou, Daquan Zhou, and Jiashi Feng.
\newblock Coordinate attention for efficient mobile network design.
\newblock {\em arXiv preprint arXiv:2103.02907}, 2021.

\bibitem{howard2019searching}
Andrew Howard, Mark Sandler, Grace Chu, Liang-Chieh Chen, Bo Chen, Mingxing
  Tan, Weijun Wang, Yukun Zhu, Ruoming Pang, Vijay Vasudevan, et~al.
\newblock Searching for mobilenetv3.
\newblock In {\em Proceedings of the IEEE/CVF International Conference on
  Computer Vision}, pages 1314--1324, 2019.

\bibitem{hu2019local}
Han Hu, Zheng Zhang, Zhenda Xie, and Stephen Lin.
\newblock Local relation networks for image recognition.
\newblock In {\em Proceedings of the IEEE/CVF International Conference on
  Computer Vision}, pages 3464--3473, 2019.

\bibitem{huang2017densely}
Gao Huang, Zhuang Liu, Laurens Van Der~Maaten, and Kilian~Q Weinberger.
\newblock Densely connected convolutional networks.
\newblock pages 4700--4708, 2017.

\bibitem{huang2018gpipe}
Yanping Huang, Youlong Cheng, Ankur Bapna, Orhan Firat, Mia~Xu Chen, Dehao
  Chen, HyoukJoong Lee, Jiquan Ngiam, Quoc~V Le, Yonghui Wu, et~al.
\newblock Gpipe: Efficient training of giant neural networks using pipeline
  parallelism.
\newblock {\em arXiv preprint arXiv:1811.06965}, 2018.

\bibitem{jiang2020convbert}
Zihang Jiang, Weihao Yu, Daquan Zhou, Yunpeng Chen, Jiashi Feng, and Shuicheng
  Yan.
\newblock Convbert: Improving bert with span-based dynamic convolution.
\newblock {\em arXiv preprint arXiv:2008.02496}, 2020.

\bibitem{krizhevsky2012imagenet}
Alex Krizhevsky, Ilya Sutskever, and Geoffrey~E Hinton.
\newblock Imagenet classification with deep convolutional neural networks.
\newblock In {\em Advances in neural information processing systems}, pages
  1097--1105, 2012.

\bibitem{lepikhin2020gshard}
Dmitry Lepikhin, HyoukJoong Lee, Yuanzhong Xu, Dehao Chen, Orhan Firat, Yanping
  Huang, Maxim Krikun, Noam Shazeer, and Zhifeng Chen.
\newblock Gshard: Scaling giant models with conditional computation and
  automatic sharding.
\newblock {\em arXiv preprint arXiv:2006.16668}, 2020.

\bibitem{liu2020improving}
Jiang-Jiang Liu, Qibin Hou, Ming-Ming Cheng, Changhu Wang, and Jiashi Feng.
\newblock Improving convolutional networks with self-calibrated convolutions.
\newblock In {\em Proceedings of the IEEE/CVF Conference on Computer Vision and
  Pattern Recognition}, pages 10096--10105, 2020.

\bibitem{liu2019roberta}
Yinhan Liu, Myle Ott, Naman Goyal, Jingfei Du, Mandar Joshi, Danqi Chen, Omer
  Levy, Mike Lewis, Luke Zettlemoyer, and Veselin Stoyanov.
\newblock Roberta: A robustly optimized bert pretraining approach.
\newblock {\em arXiv preprint arXiv:1907.11692}, 2019.

\bibitem{loshchilov2016sgdr}
Ilya Loshchilov and Frank Hutter.
\newblock Sgdr: Stochastic gradient descent with warm restarts.
\newblock {\em arXiv preprint arXiv:1608.03983}, 2016.

\bibitem{loshchilov2017decoupled}
Ilya Loshchilov and Frank Hutter.
\newblock Decoupled weight decay regularization.
\newblock {\em arXiv preprint arXiv:1711.05101}, 2017.

\bibitem{lu2019vilbert}
Jiasen Lu, Dhruv Batra, Devi Parikh, and Stefan Lee.
\newblock Vilbert: Pretraining task-agnostic visiolinguistic representations
  for vision-and-language tasks.
\newblock {\em arXiv preprint arXiv:1908.02265}, 2019.

\bibitem{paszke2019pytorch}
Adam Paszke, Sam Gross, Francisco Massa, Adam Lerer, James Bradbury, Gregory
  Chanan, Trevor Killeen, Zeming Lin, Natalia Gimelshein, Luca Antiga, et~al.
\newblock Pytorch: An imperative style, high-performance deep learning library.
\newblock {\em arXiv preprint arXiv:1912.01703}, 2019.

\bibitem{radford2019language}
Alec Radford, Jeffrey Wu, Rewon Child, David Luan, Dario Amodei, and Ilya
  Sutskever.
\newblock Language models are unsupervised multitask learners.
\newblock {\em OpenAI blog}, 1(8):9, 2019.

\bibitem{radosavovic2020designing}
Ilija Radosavovic, Raj~Prateek Kosaraju, Ross Girshick, Kaiming He, and Piotr
  Doll{\'a}r.
\newblock Designing network design spaces.
\newblock In {\em Proceedings of the IEEE/CVF Conference on Computer Vision and
  Pattern Recognition}, pages 10428--10436, 2020.

\bibitem{ramachandran2019stand}
Prajit Ramachandran, Niki Parmar, Ashish Vaswani, Irwan Bello, Anselm Levskaya,
  and Jonathon Shlens.
\newblock Stand-alone self-attention in vision models.
\newblock {\em arXiv preprint arXiv:1906.05909}, 2019.

\bibitem{rebuffi2017icarl}
Sylvestre-Alvise Rebuffi, Alexander Kolesnikov, Georg Sperl, and Christoph~H
  Lampert.
\newblock icarl: Incremental classifier and representation learning.
\newblock In {\em Proceedings of the IEEE conference on Computer Vision and
  Pattern Recognition}, pages 2001--2010, 2017.

\bibitem{simonyan2014very}
Karen Simonyan and Andrew Zisserman.
\newblock Very deep convolutional networks for large-scale image recognition.
\newblock {\em arXiv preprint arXiv:1409.1556}, 2014.

\bibitem{srinivas2021bottleneck}
Aravind Srinivas, Tsung-Yi Lin, Niki Parmar, Jonathon Shlens, Pieter Abbeel,
  and Ashish Vaswani.
\newblock Bottleneck transformers for visual recognition.
\newblock {\em arXiv preprint arXiv:2101.11605}, 2021.

\bibitem{sun2019videobert}
Chen Sun, Austin Myers, Carl Vondrick, Kevin Murphy, and Cordelia Schmid.
\newblock Videobert: A joint model for video and language representation
  learning.
\newblock In {\em Proceedings of the IEEE/CVF International Conference on
  Computer Vision}, pages 7464--7473, 2019.

\bibitem{szegedy2015going}
Christian Szegedy, Wei Liu, Yangqing Jia, Pierre Sermanet, Scott Reed, Dragomir
  Anguelov, Dumitru Erhan, Vincent Vanhoucke, and Andrew Rabinovich.
\newblock Going deeper with convolutions.
\newblock In {\em Proceedings of the IEEE conference on computer vision and
  pattern recognition}, pages 1--9, 2015.

\bibitem{szegedy2016rethinking}
Christian Szegedy, Vincent Vanhoucke, Sergey Ioffe, Jon Shlens, and Zbigniew
  Wojna.
\newblock Rethinking the inception architecture for computer vision.
\newblock In {\em Proceedings of the IEEE conference on computer vision and
  pattern recognition}, pages 2818--2826, 2016.

\bibitem{tan2019efficientnet}
Mingxing Tan and Quoc Le.
\newblock Efficientnet: Rethinking model scaling for convolutional neural
  networks.
\newblock In {\em International Conference on Machine Learning}, pages
  6105--6114. PMLR, 2019.

\bibitem{tan2019mixconv}
Mingxing Tan and Quoc~V Le.
\newblock Mixconv: Mixed depthwise convolutional kernels.
\newblock {\em arXiv preprint arXiv:1907.09595}, 2019.

\bibitem{touvron2020training}
Hugo Touvron, Matthieu Cord, Matthijs Douze, Francisco Massa, Alexandre
  Sablayrolles, and Herv{\'e} J{\'e}gou.
\newblock Training data-efficient image transformers \& distillation through
  attention.
\newblock {\em arXiv preprint arXiv:2012.12877}, 2020.

\bibitem{vaswani2017attention}
Ashish Vaswani, Noam Shazeer, Niki Parmar, Jakob Uszkoreit, Llion Jones,
  Aidan~N Gomez, Lukasz Kaiser, and Illia Polosukhin.
\newblock Attention is all you need.
\newblock {\em arXiv preprint arXiv:1706.03762}, 2017.

\bibitem{wang2020axial}
Huiyu Wang, Yukun Zhu, Bradley Green, Hartwig Adam, Alan Yuille, and
  Liang-Chieh Chen.
\newblock Axial-deeplab: Stand-alone axial-attention for panoptic segmentation.
\newblock In {\em European Conference on Computer Vision}, pages 108--126.
  Springer, 2020.

\bibitem{wang2018non}
Xiaolong Wang, Ross Girshick, Abhinav Gupta, and Kaiming He.
\newblock Non-local neural networks.
\newblock In {\em Proceedings of the IEEE conference on computer vision and
  pattern recognition}, pages 7794--7803, 2018.

\bibitem{xie2017aggregated}
Saining Xie, Ross Girshick, Piotr Doll{\'a}r, Zhuowen Tu, and Kaiming He.
\newblock Aggregated residual transformations for deep neural networks.
\newblock In {\em Proceedings of the IEEE conference on computer vision and
  pattern recognition}, pages 1492--1500, 2017.

\bibitem{yuan2021tokens}
Li Yuan, Yunpeng Chen, Tao Wang, Weihao Yu, Yujun Shi, Francis~EH Tay, Jiashi
  Feng, and Shuicheng Yan.
\newblock Tokens-to-token vit: Training vision transformers from scratch on
  imagenet.
\newblock {\em arXiv preprint arXiv:2101.11986}, 2021.

\bibitem{yun2019cutmix}
Sangdoo Yun, Dongyoon Han, Seong~Joon Oh, Sanghyuk Chun, Junsuk Choe, and
  Youngjoon Yoo.
\newblock Cutmix: Regularization strategy to train strong classifiers with
  localizable features.
\newblock In {\em Proceedings of the IEEE/CVF International Conference on
  Computer Vision}, pages 6023--6032, 2019.

\bibitem{zagoruyko2016wide}
Sergey Zagoruyko and Nikos Komodakis.
\newblock Wide residual networks.
\newblock {\em arXiv preprint arXiv:1605.07146}, 2016.

\bibitem{zhang2018network}
Daokun Zhang, Jie Yin, Xingquan Zhu, and Chengqi Zhang.
\newblock Network representation learning: A survey.
\newblock {\em IEEE transactions on Big Data}, 6(1):3--28, 2018.

\bibitem{zhang2017mixup}
Hongyi Zhang, Moustapha Cisse, Yann~N Dauphin, and David Lopez-Paz.
\newblock mixup: Beyond empirical risk minimization.
\newblock {\em arXiv preprint arXiv:1710.09412}, 2017.

\bibitem{zhang2020resnest}
Hang Zhang, Chongruo Wu, Zhongyue Zhang, Yi Zhu, Zhi Zhang, Haibin Lin, Yue
  Sun, Tong He, Jonas Mueller, R Manmatha, et~al.
\newblock Resnest: Split-attention networks.
\newblock {\em arXiv preprint arXiv:2004.08955}, 2020.

\bibitem{zhao2020exploring}
Hengshuang Zhao, Jiaya Jia, and Vladlen Koltun.
\newblock Exploring self-attention for image recognition.
\newblock In {\em Proceedings of the IEEE/CVF Conference on Computer Vision and
  Pattern Recognition}, pages 10076--10085, 2020.

\bibitem{zhao2020point}
Hengshuang Zhao, Li Jiang, Jiaya Jia, Philip Torr, and Vladlen Koltun.
\newblock Point transformer.
\newblock {\em arXiv preprint arXiv:2012.09164}, 2020.

\bibitem{zheng2020rethinking}
Sixiao Zheng, Jiachen Lu, Hengshuang Zhao, Xiatian Zhu, Zekun Luo, Yabiao Wang,
  Yanwei Fu, Jianfeng Feng, Tao Xiang, Philip~HS Torr, et~al.
\newblock Rethinking semantic segmentation from a sequence-to-sequence
  perspective with transformers.
\newblock {\em arXiv preprint arXiv:2012.15840}, 2020.

\bibitem{zhong2020random}
Zhun Zhong, Liang Zheng, Guoliang Kang, Shaozi Li, and Yi Yang.
\newblock Random erasing data augmentation.
\newblock In {\em Proceedings of the AAAI Conference on Artificial
  Intelligence}, volume~34, pages 13001--13008, 2020.

\bibitem{zhou2020rethinking}
Daquan Zhou, Qibin Hou, Yunpeng Chen, Jiashi Feng, and Shuicheng Yan.
\newblock Rethinking bottleneck structure for efficient mobile network design.
\newblock {\em ECCV, August}, 2, 2020.

\end{thebibliography}
}

\newpage\null\thispagestyle{empty}

\appendix

\section{Experiment Implementation Details}
\label{supp:impl_details}

\paragraph{Attention reuse:} As shown in Fig. 3(b) and Tab. 3 in the main paper, the vision transformers with 24 blocks and 32 blocks have 11 and 15 blocks with similar attention maps, respectively. 
To verify the effectiveness of the attention maps from those blocks, 
we directly force those blocks to share the same attention map 
as the last `unique' block as defined in Sec. 5.2. 
Specifically, we take the attention map of the last `unique' block and 
denote it as $\mathbf{A}_{unique}$.
For all the following blocks, the attention output is calculated by:
\begin{equation}
    \label{eqn:supp_share_block}
    \text{Attention}(Q,K,V) = \text{Norm}(\Theta^\top \mathbf{A}_{unique})V,
\end{equation}
where $\Theta$ is used to simulate the small variance between attention maps across layers since they are not identical. Norm is batch normalization used to adjust the variance across layers. As shown in Tab. 3, for a ViT with 32 blocks, forcing the top 15 blocks to share the same attention map causes negligible degradation on the classification accuracy on ImageNet. This proves that adding those blocks do not contribute to the accuracy improvement.

\paragraph{Training loss:} We use the cross-entropy (CE) loss as the training loss for all experiments. To minimize the similarity of the attention maps across layers, we add the cosine similarity between layers into the loss function when training the model. 
 
 \begin{equation}
    \label{eqn:loss_fn}
    \text{Loss}_{train} = \text{Loss}_{CE} + \lambda \sum_{l=0}^{B} \text{Similarity}(\mathbf{A}^l, \mathbf{A}^{l+1})
\end{equation}
where $\text{Similarity}(\mathbf{A}^l, \mathbf{A}^{l+1})$ denotes the cosine similarity between layer $l$ and $l+1$ and $\mathbf{A}^l$ denotes the attention map of layer $l$. B denotes the number of bottom blocks used for regularization and is a hyper-parameter. We set B to 4, 8 and 12 for training ViT models with 16, 24 and 32 blocks respectively.

\begin{figure}[t]
\begin{center}
\includegraphics[width=0.99\linewidth]{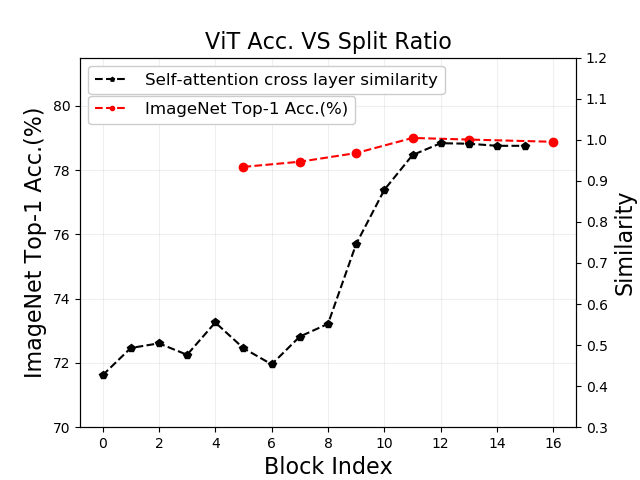}
\end{center}
  \caption{ViT classification accuracy with \nameofatten{} applied on different number of blocks. The black dotted line denotes the cosine similarity ratio between adjacent blocks of the original ViT model with 16 blocks. The red dotted line denotes the top-1 classification accuracy on ImageNet. The accuracy of the model with blocks index k denotes that the \nameofatten{} is applied on top $(16-k)$ blocks.}
\label{fig:supp_split_ratio}
\end{figure}

\begin{table}[h]
\footnotesize
\caption{Structural hyper-parameter of DeepViT-S and DeepViT-L. Note that the embedding dimension is slightly larger than the baseline models. This is to adjust the size of the model to have a comparable size with other methods for a fair comparison.}
\label{tab:supp_sota_comparison_arch}
% \begin{center}
\centering
\begin{tabular}{lccccc}
\toprule
% \multicolumn{1}{c}{\bf Group}  
\bf Model
% & \bf Searching Algo.
&\bf \#Blocks
&\bf \#Embedding
% &\bf \#Heads
&\bf MLP size
&\bf Split ratio
% &\bf Param.(M)
% &\bf MAdds (M)
% &\bf Acc. (Top-1 \%)
\\ \midrule %\\
% \multirow{10}{*}{\makecell[l]{w/o SE}} 
% & MBV2-0.75 &- & 2.6 & 209 & 69.8 \\
 DeepViT-S & 16 & 396 & 1188  & 11-5 \\
 DeepViT-L &  32 & 420 & 1260  & 20-12 \\ 
% &Ours$^*$(Voted)& Gradient + Manual &  5.22 & 420 & 77.88\\
\bottomrule
\vspace{-5mm}
\end{tabular}
% \end{center}
\end{table}

\section{DeepViT architecture design}
As observed in Fig. 3(a), the attention maps of the transformer blocks become similar only at the top blocks. Thus, it is not necessary to apply re-attention to all blocks. To study the optimal number of blocks with re-attention, we conduct a set of experiments on a ViT model with 16 transformer blocks. For each experiment, we only apply re-attention on the top $K$ blocks where $K$ ranges from 5 to 15. The rest of the blocks are using the original transformer block structure. We train each model on ImageNet with the same set of training hyper-parameters as those for baseline models as detailed in Sec. 5 in the main paper. The results are shown in Fig. \ref{fig:supp_split_ratio}. 

It is observed that, as the number of re-attention blocks varies, the top-1 classification accuracy changes correspondingly. The highest accuracy appears at the position where the number of re-attention blocks is the same as the number of similar attention map blocks. Based on this observation, we define the architecture of DeepViT-S and DeepViT-L with 5 and 12 re-attention blocks respectively. Detailed configurations are shown in Tab. \ref{tab:supp_sota_comparison_arch}. Note that we adjust the embedding dimension to have a comparable size with other methods.

\section{Impacts of hyper-parameters}
In the main paper, all experiments are run with the same set of training hyper-parameters as the one used for reproducing ViT models.
However, as shown in \cite{touvron2020training}, an improved training recipe could improve the performance of ViT models significantly. In Tab. \ref{tab:supp_comaprison_deit}, we present the performance of DeepViT-S and DeepViT-L with the same set of training recipes as DeiT except that we do not use repeated augmentation. 
In Tab. \ref{tab:supp_comaprison_deit}, it is clearly shown that the performance of DeepViT could be further improved with optimized training hyper-parameters. 

\begin{table}[h]
\footnotesize
\caption{DeepViT model with different training recipes. $^{\star}$ denotes the model trained with DeiT \cite{touvron2020training} training recipes. }
\label{tab:supp_comaprison_deit}
% \begin{center}
\centering
\begin{tabular}{lcccc}
\toprule
% \multicolumn{1}{c}{\bf Group}  
\bf Model
% & \bf Searching Algo.
&\bf Params. (M)
&\bf MAdds (G)
&\bf Acc. (\%)
\\ \midrule %\\
 DeiT-S \cite{touvron2020training} & 22 & -- & 79.8 \\
 DeiT-S (KD) \cite{touvron2020training} & 22 & -- & 81.2 \\
%  \midrule
  DeepVit-S (Ours) & 27  & 6.2 & 81.4 \\ 
  DeepVit-S$^{\star}$ (Ours) & 27  & 6.2 & 82.3 \\ 
 \midrule
 DeiT-B \cite{touvron2020training} & 86 & 17.7 & 81.8 \\
 DeiT-B (KD) \cite{touvron2020training} & 86 & 17.7 & 83.4 \\
%   \midrule
 DeepViT-L (Ours) & 55  & 12.5 & 82.2 \\ 
 DeepViT-L$^{\star}$ (Ours) & 58  & 12.8 & 83.1 \\ 
\bottomrule
\vspace{-5mm}
\end{tabular}
% \end{center}
\end{table}

\section{Similarity calculation}

\myPara{Cosine similarity between layers} To measure the similarity between the attention maps, we define the similarity $S^{p,q}$ between the attention maps of two layers, $p$ and $q$, as the ratio of the number of similar vector pairs to the total number of pairs between two attention maps:
\begin{equation}
    S(p,q) = \frac{\sum I}{|M^{p,q}|}, \quad I_{i,j} =
    \begin{cases}
      1, & \text{\quad if $M^{p,q}_{i,j}$ $>$ $\tau$}\\
      0, & \text{\quad otherwise}
    \end{cases}
\end{equation}
where $\tau$ is a hyper-parameter and used as a threshold for deciding similar vectors\footnote{0.5 is selected as a threshold for visualization purpose in this paper}. 

\myPara{Definition of similar blocks} A block is counted as a similar block if the similarity between it's attention map and the adjacent block's attention map is larger than 80\%. To measure the block similarity for a ViT model with $B$ blocks, we take the ratio of the number of similar blocks to the total number of blocks as a measurement.

\end{document}